\documentclass{article}

\usepackage[main, final]{neurips_2026}

\usepackage[utf8]{inputenc} % allow utf-8 input
\usepackage[T1]{fontenc}    % use 8-bit T1 fonts
\usepackage{hyperref}       % hyperlinks
\usepackage{url}            % simple URL typesetting
\usepackage{booktabs}       % professional-quality tables
\usepackage{amsfonts}       % blackboard math symbols
\usepackage{nicefrac}       % compact symbols for 1/2, etc.
\usepackage{microtype}      % microtypography
\usepackage[dvipsnames]{xcolor}         % colors
\usepackage[normalem]{ulem} % strike-through \sout

\usepackage{mycommands}

\usepackage{placeins}
\usepackage{fontawesome5}  % \faGithub next to the code link under the authors

\usepackage{comment}
\usepackage{tcolorbox}
\usepackage[cachedir=minted-cache]{minted}
\tcbuselibrary{listings,minted}
\usepackage{tcolorbox}                        
\tcbuselibrary{minted, skins, breakable}                                                                                                                  \AtBeginEnvironment{minted}{\setlength{\parskip}{0pt}\setlength{\partopsep}{0pt}\setlength{\topsep}{0pt}}
\fvset{baselinestretch=1, samepage=true}                   
\newtcblisting{repl}{%
  listing engine=minted,%
  minted language=python,%
  minted options={fontsize=\footnotesize, breaklines=true, baselinestretch=1, xleftmargin=0pt, xrightmargin=0pt, escapeinside=||},%
  colback=orange!8, colframe=orange!8,%
  enhanced, frame hidden, arc=2pt,%
  boxsep=0pt,%
  left=0pt, right=0pt, top=0pt, bottom=0pt,%
  before skip=2pt, after skip=2pt,%
  listing only,%
  before upper={{\ttfamily\footnotesize\color{blue!50!black}<repl>}\par},%
  after upper={\par{\ttfamily\footnotesize\color{blue!50!black}</repl>}},%
}

\newtcblisting{observation}{%
  listing engine=minted,%
  minted language=text,%
  minted options={fontsize=\footnotesize, breaklines=true, baselinestretch=1, xleftmargin=0pt, xrightmargin=0pt, escapeinside=||},%
  colback=orange!8, colframe=orange!8,%
  enhanced, frame hidden, arc=2pt,%
  boxsep=0pt,%
  left=0pt, right=0pt, top=0pt, bottom=0pt,%
  before skip=2pt, after skip=2pt,%
  listing only,%
  before upper={{\ttfamily\footnotesize\color{gray!50!black}<observation>}\par},%
  after upper={\par{\ttfamily\footnotesize\color{gray!50!black}</observation>}},%
}

\newtcolorbox{namespacebox}[1]{%
  enhanced, breakable,%
  colback=orange!8, colframe=orange!8,%
  frame hidden, arc=2pt,%
  boxsep=0pt,%
  coltitle=orange!30!black,%
  fonttitle=\bfseries\ttfamily\small,%
  title={namespace: #1},%
  left=6pt, right=0pt, top=4pt, bottom=4pt,%
  before upper={\color{orange!40!black}\ttfamily\footnotesize},%
}
\newcommand{\task}[1]{{\color{black}\textbf{Task:} #1}}

\usepackage{multirow}
\usepackage{diagbox}
\usepackage{makecell}
\usepackage{slashbox}
\usepackage{wrapfig}

\usepackage{lettrine}

\usepackage{svg}

\usepackage[scr]{rsfso}  % for mathsrc

\usepackage{bm}        % for \bm
\usepackage{enumitem} % for enumeration tweaks
\AtBeginEnvironment{repl}{\setlength{\parskip}{0pt}\setlength{\partopsep}{0pt}\setlength{\topsep}{0pt}}
\AtBeginEnvironment{observation}{\setlength{\parskip}{0pt}\setlength{\partopsep}{0pt}\setlength{\topsep}{0pt}}

\tcbset{
  taskbox/.style={
    sharp corners, fontupper=\small,
    colback=white, colframe=black!25, boxrule=0.6pt,
    title={\textbf{\small Task:}},
    colbacktitle=white, coltitle=black, titlerule=0pt
  },
  shallowbox/.style={
    sharp corners, fontupper=\small,
    colback=oc-yellow-0, colframe=oc-yellow-8, boxrule=0.6pt,
    title={\textbf{\small Incremental Decomposition:}},
    colbacktitle=oc-yellow-0, coltitle=black, titlerule=0pt
  },
  codebox/.style={
    sharp corners,
    colback=black!5, colframe=black!35, boxrule=0.6pt,
    title={\textbf{\small Code:}},
    colbacktitle=black!5, coltitle=black, titlerule=0pt,
    listing engine=minted, minted language=python, listing only
  }
}
\newtcblisting{codeblock}[1][]{
  codebox,
  minted options={fontsize=\small, linenos, numbersep=3mm, breaklines, #1}
}
\usepackage{graphicx}
\usepackage{subcaption}

\usepackage{amsmath}
\usepackage{amssymb}
\usepackage{longtable}
\usepackage{tikz}

\newcommand{\openquote}{%
  \makebox[0.6em][l]{\smash{\raisebox{-0.4ex}{\large ``}}}%
}
\newcommand{\closequote}{%
  \unskip
  \makebox[0.6em][r]{\smash{\raisebox{-0.4ex}{\large ''}}}%
}
\definecolor{forestgreen}{rgb}{0.13,0.55,0.13}

\definecolor{OliveGreen}{HTML}{156e2d}
\newcommand*\circled[1]{\tikz[baseline=(char.base)]{
            \node[shape=circle,draw,inner sep=0.5pt] (char) {#1};}} 

\newcommand{\MRL}{\textrm{OnION}}
\newcommand{\DR}{\textsc{Dolores}}
\newcommand{\DRb}{\textsc{\textbf{D}o\textbf{l}o\textbf{r}e\textbf{s}}}
\newcommand{\Ical}{\ensuremath{\textcolor{RawSienna}{\mathcal{I}}}}

\newcommand{\Ecal}{\ensuremath{\textcolor{blue}{\mathcal{E}}}}
\newcommand{\Acal}{\ensuremath{\textcolor{WildStrawberry}{\mathcal{A}}}}
\newcommand{\Malone}[1]{\ensuremath{\textcolor{OliveGreen}{\mathcal{M}}}}

\newcommand{\MFmod}[2]{\textcolor{OliveGreen}{\mathcal{M}_{#1}(}#2\textcolor{OliveGreen}{)}}
\newcommand{\IDeep}{\ensuremath{\textcolor{RawSienna}{\mathcal{I}_{\mathrm{D}}}}}
\newcommand{\mhl}{\textcolor{OliveGreen}{m}}
\newcommand{\Ifn}[1]{\ensuremath{\textcolor{RawSienna}{\mathcal{I}(}\openquote{}\text{#1}\closequote{}\textcolor{RawSienna}{)}}}
\newcommand{\Afn}[1]{\ensuremath{\textcolor{WildStrawberry}{\mathcal{A}(}\openquote{}\text{#1}\closequote{}\textcolor{WildStrawberry}{)}}}
\newcommand{\Efn}[2]{\ensuremath{\textcolor{blue}{\mathcal{E}}_{#1}\textcolor{blue}{(}\openquote{}\text{#2}\closequote{}\textcolor{blue}{)}}}
\newcommand{\mfn}[2]{\ensuremath{\mhl_{#1}\textcolor{OliveGreen}{(}\openquote\text{#2}\closequote{}\textcolor{OliveGreen}{)}}}
\newcommand{\indii}{\hspace*{2em}}

\newcommand{\IDo}{\ensuremath{\textcolor{RawSienna}{\IDeep(}}}
\newcommand{\IDc}{\ensuremath{\textcolor{RawSienna}{)}}}
\newcommand{\Io}{\ensuremath{\textcolor{RawSienna}{\Ical(}}}
\newcommand{\Ic}{\ensuremath{\textcolor{RawSienna}{)}}}

\newcommand{\Eo}{\ensuremath{\textcolor{blue}{\Ecal(}}}
\newcommand{\Ec}{\ensuremath{\textcolor{blue}{)}}}
\newcommand{\Ao}{\textcolor{WildStrawberry}{\ensuremath{\Acal(}}}
\newcommand{\Ac}{\textcolor{WildStrawberry}{\ensuremath{)}}}

\newcommand{\bsep}{\textcolor{blue}{\textbf{;}}}

\tcbset{ebox/.style={sharp corners, fontupper=\small\ttfamily,
  left=3pt, right=3pt, top=1pt, bottom=1pt,
  colback=blue!8, colframe=blue!35, boxrule=0.5pt}}

\usepackage{subcaption}

\usepackage{lineno}

\definecolor{darkblue}{rgb}{0, 0, 0.5}
\newlength{\DRsecskip}    \DRsecskip    =0pt    % after \section
\newlength{\DRsubsecskip} \DRsubsecskip =0pt    % after \subsection
\newlength{\DRparvskip}   \DRparvskip   =0pt    % before \paragraph

\newcommand{\secvs}{\vspace{\DRsecskip}}
\newcommand{\subsecvs}{\vspace{\DRsubsecskip}}
\newcommand{\parvs}{\vspace{\DRparvskip}}

\title{Deep Reasoning in General Purpose Agents via Structured Meta-Cognition}

\author{
  \textbf{Dean Light}$^1$\thanks{Equal contribution.} \quad
  \textbf{Michael Theologitis}$^1$\footnotemark[1] \quad 
  \textbf{Kshitish Ghate}$^{1}$\footnotemark[1] \\
  \textbf{Shuyue Stella Li}$^1$ \quad 
  \textbf{Benjamin Newman}$^1$ \quad 
  \textbf{Chirag Shah}$^1$ \quad 
  \textbf{Aylin Caliskan}$^1$ \\
  \textbf{Pang Wei Koh}$^1$ \quad 
  \textbf{Dan Suciu}$^1$ \quad 
  \textbf{Yulia Tsvetkov}$^1$ \\
  $^1$University of Washington \\
  Seattle, WA, USA \\
  \texttt{\{deanlcs, mthe, kghate\}@cs.washington.edu} \\[0.5em]
  \faGithub~\codelink
}

\newif\ifDRacks      \DRackstrue
\newif\ifDRchecklist \DRchecklisttrue
\newcommand{\DRvenuestatements}{}
\newcommand{\DRendmainmatter}{}
\newcommand{\DRcompetinginterests}{}

\newcommand{\codelink}{\url{https://github.com/DeanLight/dolores}}

\DRackstrue

\DRchecklisttrue

\renewcommand{\DRendmainmatter}{\clearpage}

\renewcommand{\DRcompetinginterests}{%
  \par The authors declare no competing interests.%
}

\begin{document}

% ===========================================================================
% SHARED BODY -- the text of the paper.  Identical for every version;
% nothing in here is style- or venue-specific.
%
% Version-dependent pieces are reached through macros/toggles that the driver
% files define (see preamble.tex):
%   \codelink       -- code repository URL (anonymized vs. public)
%   \ifDRacks       -- Acknowledgements section on/off
%   \ifDRchecklist  -- NeurIPS paper checklist on/off (the two NeurIPS versions)
%   \DRvenuestatements -- hook expanded just before \bibliography
%   \DRendmainmatter   -- hook expanded right after the conclusions
%   \DRcompetinginterests -- appended to the acknowledgements
% ===========================================================================

% \ifcolmsubmission
% \linenumbers
% \fi

\maketitle

\begin{abstract}
Humans solve complex problems by flexibly shifting among reasoning modes, often without explicit deliberation: they plan, execute, revise intermediate goals, resolve ambiguity through associative judgment, and apply formal procedures to well-specified subproblems. Current LLM agents lack this flexibility, as their scaffolds hard-code such reasoning decisions in advance through fixed inference patterns. These scaffolds are effective when their prescribed structure matches the task, but brittle when solving the task requires adapting the structure of reasoning itself. We introduce \textbf{Deep Reasoning -- an inference-time approach for constructing task-specific scaffolds through structured meta-reasoning}. Deep Reasoning uses a formal language that represents meta-reasoning as executable decompositions over associative inference, formal computation, and recursive subproblem solving, enabling decomposition principles to be encoded as in-context examples that guide test-time scaffold construction. We instantiate this approach in a general-purpose agent (\DR{}) that distributes complex tasks across smaller, more controlled reasoning threads while preserving dependencies among subproblems. We evaluate \DR{} against state-of-the-art scaffolding methods across four hard benchmarks: grounded multi-hop reasoning, synthetic long-chain question answering, long-context aggregation, and deep research-style information seeking. \DR{} outperforms all evaluated scaffolds across four benchmarks, three model sizes, and two model families, improving over the strongest evaluated scaffold baseline by 24.8\% on average, including methods tailored to individual benchmark families. Trace and token analyses suggest that while baseline scaffolds fail by overloading individual LLM calls, \DR{} succeeds by distributing cognition across structured, lower-load reasoning threads, thereby reducing premature termination and hallucination. This advantage can even bridge the scaling gap, with an 8B version surpassing all evaluated 32B baselines from the same family in more than half the settings. 
These results point toward future agentic systems that treat scaffolding as adaptive reasoning, constructing the structure each task requires just-in-time.
\end{abstract}

\begin{comment}

New format

Intro
 AI REASONING
 Current Advances in AI reasoining
Background - CogSci
DeepReasoning
  Theoretical Framework
  Simple implementation
Results
  Expressivity
  Empirical evaluations
    Setup
    Figures

\end{comment}

\section{Introduction}\label{sec:intro}

Humans intuitively solve complex problems through meta-reasoning, a cognitive process in which we model the task and environment, select appropriate problem-solving strategies, break problems into manageable steps, plan how to solve them, and execute these plans while tracking intermediate reasoning and goals~\citep{newell1959report, flavell1979metacognition}. Crucially, meta-reasoning allows us to integrate multiple modes of cognition, including formal reasoning, based on explicit rules and logical operations, and associative reasoning, which leverages intuition and patterns from prior experience~\citep{bellini2022dual}. The interplay between these two is dynamic and task-dependent, enabling us to be effective problem-solvers across a wide range of domains, including but not limited to mathematics, information synthesis, programming and creative writing~\citep{stanovich2011rationality}. 
% \bn{nit: it feels awkward to describe humans as being good at ``multi-hop qa'' or ``code generation'' as opposed to ``information synthesis'' and ``software engineering''.}\dl{good point, ammended}

\begin{figure}[t]
    \centering
\includegraphics[width=0.99\textwidth]{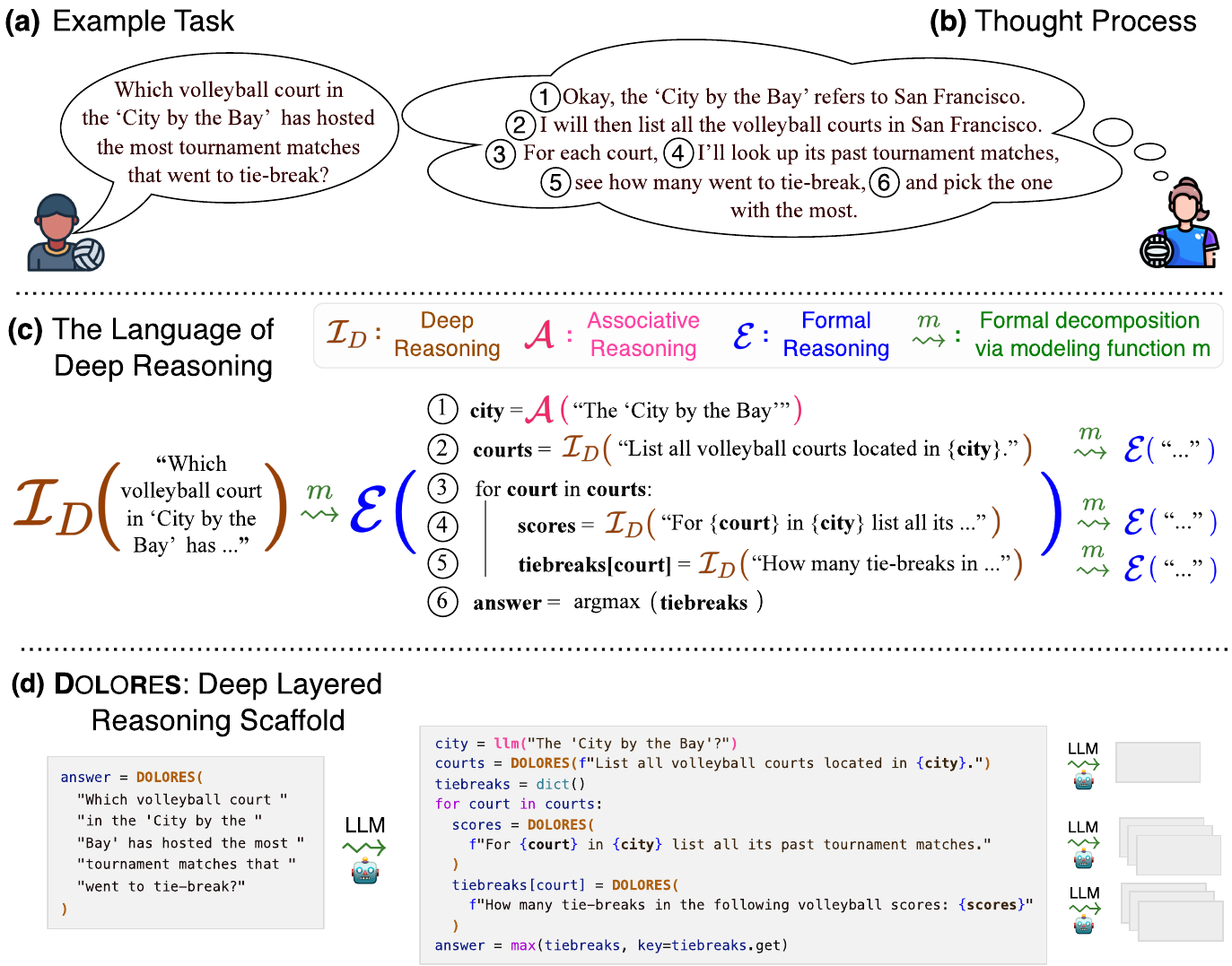}
    \caption{Deep Reasoning leverages human meta-reasoning traces to build ``just-in-time'' scaffolds. (\textbf{a}): A task is given to a human. (\textbf{b}):~The human intuitively meta-reasons about how to solve it. (\textbf{c}):~These verbalized traces are then directly mapped into the language of Deep Reasoning. (\textbf{d}):~A Deep Reasoning agent like \DR{} (\S\ref{sec:deep_reasoning}) uses them as in-context examples to guide its own meta-reasoning abilities. Example \textit{adapted} from the DeepSearchQA benchmark~\citep{gupta2026deepsearchqa}.
    \vspace{-10pt}
    }
    \label{fig:conceptual_onion}
\end{figure}

% \bn{Nit: The top line of (c) looks a bit like an equation, which confused me at first bc of minus sign. Maybe can put light gray boxes behind the $I_D$, $A$, etc to visually separate them?}

%\bn{I'd add a few citations to this para.}
% \kg{connect why we need LLM reasoning? This seems to start abruptly}s

Attempting to mimic human reasoning, LLMs achieve impressive results across a variety of both creative (e.g., writing) and formal (e.g., programming and mathematics) domains. However, there is growing evidence that this is ``shallow reasoning'', a surface-level imitation of verbalized human reasoning that does not reliably track underlying thought processes~\citep{kargupta2025cognitive}. In particular, LLMs struggle to reason formally in a robust way, which is in line with work showing LLMs to be unfaithful to their own reasoning traces~\citep{lanham2023measuring, yee2024dissociation, lyu2023faithful, arcuschin2025chain} and to degrade in performance on long or cognitively demanding reasoning tasks~\citep{jaech2024openai, chen2025not, hassid2025don}.

% \kg{humans intuitively solve complex problems through meta reasoning}

To address these limitations, progress in LLM-based systems has increasingly relied on carefully designed agentic scaffolds~\citep{YangJWLYNP24, DBLP:journals/corr/abs-2511-03690, smolagents}. These scaffolds are programs that coordinate communication between LLMs and external tools~\citep{rosser2026agentbreeder}. However, they typically \textit{predefine} how a task should be decomposed, and how the reasoning should be carried out. For example, ReAct~\citep{yao2022react} breaks down the task in a sequence of associative reasoning steps with tool calls. CodeAct~\citep{wang2024executable} replaces these tool calls with formal reasoning via code. Deep Research~\citep{roucher_open_deep_research_2025} prescribes a two-layer decomposition, where a manager agent delegates subproblems to search agents. Finally, RLMs~\citep{zhang2025recursive} decompose long-input tasks by recursively chunking them inside a programmatic environment via recursive self-calls. The common pattern across these scaffolds is that they hard-code how to decompose reasoning. Essentially, they fix how to meta-reason about a task without seeing it, and, as a result, fail when other reasoning behaviors are required~\citep{fu2025agentrefine}.

In this work, we leverage insights from  human meta-reasoning to build ``just-in-time'' meta-reasoning scaffolds. While humans are able to meta-reason intuitively, there is no formal way capture these processes and translate them for agentic systems. To this end, we ground our approach in the cognitive science literature (\S\ref{sec:cogsci}) and introduce the language of Deep Reasoning (\S\ref{sec:theory}), a formal language that combines associative and formal reasoning through structured meta-reasoning. Deep Reasoning provides a principled way to take human meta-reasoning traces, formalize them, and convert them into atomic decompositions usable by agentic systems in-context. It allows us to instill how humans decompose and reason about tasks into agents just-in-time (Figure~\ref{fig:conceptual_onion}).

% In this work, we leverage human meta-reasoning to build ``just-in-time'' meta-reasoning scaffolds. While humans are able to meta-reason intuitively, there is no formal way capture these processes and translate them for agentic systems. To this end, we first propose \MRL{} (\S\ref{sec:theory}), a formal language that combines associative and formal reasoning through structured meta-reasoning, grounded in the cognitive science literature (\S\ref{sec:cogsci}). We introduce a principled way to take human meta-reasoning traces, formalize them in \MRL, and convert them into atomic decompositions usable by agentic systems in-context~(\S\ref{sec:design-principles}). This enables us to instill how humans decompose and reason about tasks into agents in-context.

% \mt{narrative of sections}

% This conversion is guided by three design principles of how reasoning should be broken down (\S\ref{sec:design-principles})

% We show how to take human meta-reasoning traces, formalize them in \MRL, and then directly convert them into atomic decompositions usable by agentic systems in-context following three design principles (\S\ref{sec:design-principles}). This enables us to instill how humans decompose and reason about tasks into agentic systems on the fly.

% It outperforms state-of-the-art methods by significant margins, achieving an average improvement of 24.8\% across four reasoning tasks, three model sizes, and two model families.\mt{more specific - get to the point}

%\bn{nit: good practice to spell out acronym first time it's mentioned}

We instantiate this approach in an agent we call \DRb{} (Deep Layered Reasoning Scaffold, \S\ref{sec:deep_reasoning}), which uses in-context meta-reasoning examples to guide its reasoning and dynamically adapt its scaffold at test time. It outperforms all evaluated state-of-the-art scaffold methods, with an \textbf{average improvement of 24.8\% over the best-performing baseline across four reasoning tasks}, three model sizes, and two model families. Notably, we find that \DR{} equipped with an 8B model outperforms all baselines that use a model from the next scaling tier (32B) of same family in more than half of the evaluated settings. By analyzing reasoning traces, we also show that all other scaffolds tend to hallucinate and terminate prematurely as they delegate too much to a single LLM context thread. In contrast, \DR{} avoids these issues by structuring reasoning in more fine-grained ways, decomposing it into \textit{atomic} associative, formal, or meta-reasoning steps that can be handled reliably by a single LLM memory thread. 

Our contributions are twofold: \circled{1} the language of Deep Reasoning, a formal language for capturing human meta-reasoning traces in a way that can be used by agentic systems, and \circled{2} \DR, an agent that operationalizes our approach and demonstrates improved performance against state-of-the-art methods. Our code is available at \codelink.

\section{Background}\label{sec:cogsci}

% \mt{MT -- todo. connect to prev well.}
% This section introduces concepts and prior work from the cognitive science and AI literature on which \MRL{} is based.
We build the Deep Reasoning language, a formal language for capturing human meta-reasoning, on concepts from prior work in cognitive science, which are discussed below.

% \bn{\sout{artificial intelligence} nit: Reads a bit awkwardly to me, I think bc building on AI work is assumed. (pls feel free to ignore)}

% \bn{explicitly saying ``literature'' sounds off to me. As an alternative: ``We base our formal language for meta-reasoning, \MRL, on concepts from prior work in cognitive science and artificial intelligence, which are discussed below.''} - Done

\subsection{Axes of Reasoning}\label{sec:reasoning_types}

\subsecvs
Prior work characterizes reasoning along two orthogonal dimensions: \textit{how} reasoning is carried out and what the reasoning is \textit{about}. To make these distinctions concrete, we refer throughout to the running example introduced in Figure~\ref{fig:conceptual_onion} about the volleyball court.
%\stella{about the volleyball court}.
% \pw{emphasize `what' instead of `about' for consistency with `how'}

\begin{wrapfigure}[14]{r}{0.37\textwidth}
  \centering
  \vspace{-5mm}
  \includegraphics[width=\linewidth]{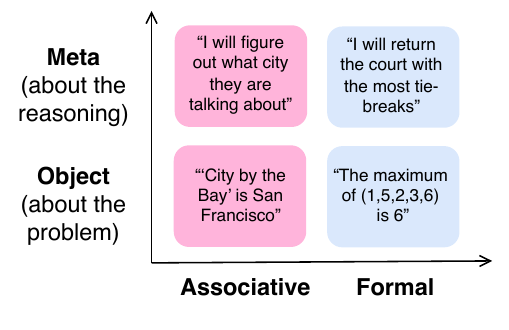}\vspace{-1.5mm}
  \caption{Associative and formal reasoning describe \emph{how} reasoning is carried out, while meta and object levels describe what the reasoning is \textit{about}.}
  \label{fig:axes-of-reasoning}
\end{wrapfigure}

% \bn{Something that initially confused me was that I was interpreting the top right corner as ``reasoning formally about how the reasoning is carried out'', rather than ``reasoning about how to do formal reasoning.'' The examples are super helpful in clearing this up!} \mt{thank you ben :)}

\parvs
\paragraph{Associative vs. Formal (how).} When solving a task, some steps rely on intuition and associations, while others follow explicit rules. Associative reasoning generally operates through intuitive proximity shaped by memory and context \citep{mednick1962associative, sloman1996empirical}. For example, interpreting ``The City by the Bay'' as San Francisco can be done associatively drawing on experience and abstract links between concepts. In contrast, formal reasoning proceeds via the sequential application of logical rules over structured, well-defined representations \citep{wason1968reasoning, rips1983cognitive, johnson1991deduction}. For example, counting the number of volleyball games that went to a tie-break and selecting the maximum follows a clear, rule-based procedure, and is therefore formal reasoning.
% \yt{i wasn't sure if that's completely comprehensive characterization of axes of reasoning. for example, i remember older papers from yejin on "abductive reasoning" -- does this type of reasoning fall under associative of deductive? no need to explicitly address this comment in writing it's more a question for my own clarification}
% \yt{\emph{what} instead of \emph{about}?; also pink and blue squares feel a bit like a waste of space, can we put some content there?}

\parvs
\paragraph{Object vs. Meta-level (about what).} 
At the same time, some reasoning steps operate on the problem itself, while others organize how to solve it. Object-level reasoning operates on objects in the task. For example, interpreting the ``City by the Bay'' and counting the number of tie-breaks are object-level reasoning, since they act directly on the task. In contrast, meta-level reasoning operates over the reasoning process: selecting strategies, decomposing problems, and guiding inference by constructing and updating an internal model of the task, state, and goals \citep{newell1959report, flavell1979metacognition, ackerman2017meta}. Deciding to break a task into steps (e.g., ``first identify the city, then list volleyball courts, then count...''), without executing the actual steps, is meta-level reasoning. 
%
% \bn{I understand this distinction, but it feels more fuzzy than the associative-deductive distinction} \pw{I agree: deciding that I first need to list volleyball courts and then count seems more like `reasoning' than `meta-reasoning' to me?}.
% \stella{Meta-reasoning is more about the awareness that ``I need to have a plan'' ``I need to pick which strategy is better based on the nature of the problem which is XX'' or ``I need t%o decompose this''}
%
These two dimensions are orthogonal (Figure~\ref{fig:axes-of-reasoning}). For example, we can have associative meta-reasoning or formal object-level reasoning.

% At the same time, some reasoning steps operate on the problem itself, while others organize how to solve it. Interpreting the ``City by the Bay'' and counting the number of tie-breaks are object-level reasoning, since they act directly on the task. More generally, object-level reasoning concerns inference within the problem domain. In contrast, breaking the task into steps (e.g., ``first identify the city, then list volleyball courts, then count...'') is meta-reasoning, since it structures the solution process. Meta-reasoning operates over the reasoning process itself: selecting strategies, decomposing problems, and guiding inference by constructing and updating an internal model of the task, state, and goals \citep{newell1959report, flavell1979metacognition, ackerman2017meta} \bn{I understand this distinction, but it feels more fuzzy than the associative-deductive distinction} \pw{I agree: deciding that I first need to list volleyball courts and then count seems more like `reasoning' than `meta-reasoning' to me?}.

% These two dimensions are orthogonal. For example, we can have associative meta-reasoning or formal object-level reasoning.
% % \bn{probably not a priority, but if it's possible to fit examples into each box, it would be very helpful.}

\subsection{Human Reasoning} 
% \bn{I think you can move this section to the appendix, and just append the first sentence to the end of the previous section.}\dl{We need this section in the main text to ground the need for translation from our capacities to LLM capacities} 
\subsecvs
Humans tightly interleave meta- and object-level reasoning when solving tasks that go beyond their cognitive capacity (total processing resources available in working memory). When working through a problem, we search for structure, decompose it into manageable subproblems, solve them, and integrate the results according to that structure~\citep{GRIFFITHS201924,Sweller2019}. This search for structure is a form of meta-reasoning, often referred to as ``modeling'' or ``mental modeling''. Our strong reasoning and generalization abilities are often attributed to these meta-reasoning and modeling capabilities~\citep{kargupta2025cognitive}. These flexible reasoning abilities are also deeply rooted in our use of natural language. Language plays a dual role in human reasoning: we sometimes use it formally (e.g., ``How many apples are in the basket?''), while other times we use it associatively (e.g., ``Is this apple ripe?''). Moreover, when solving complex problems we use natural language to interleave associative reasoning and formal reasoning, at both the object and meta levels~\citep{broadbent1958perception,treisman1964monitoring,Rosch1978-ROSPOC-8}. It is this flexible, structure-dependent interleaving that enables us to generalize so effectively across tasks~\citep{stanovich2011rationality, bellini2022dual}.

% The principles also trade off against one another: aggressive symbolic delegation typically demands deeper meta-reasoning to find the right formal handle, and over-decomposition adds overhead when a task already fits within the LLM's cognitive capacity.

\section{Deep Reasoning -- A Formal Language for Meta-Reasoning}\label{sec:theory}
\secvs

\subsection{Motivation}\label{sec:design-principles}
% \subsection{Motivation}
% \bn{I feel like a definition of cognitive capacity could be helpful, though I have an intuitive understanding of what is meant}\dl{I added defintion in parenths in prev section. Does it work now?}
\subsecvs
LLMs, despite their large context windows, have a limited cognitive capacity~\citep{DBLP:conf/emnlp/ZhangJOV24, DBLP:conf/nips/Chen0WZC24, DBLP:conf/acl/FuQH0ZL25}. However, their cognitive capacity is substantially different from that of humans. For example, they are able to easily memorize and retrieve large collections of text and code while still struggling in basic reasoning tasks~\citep{deepmind25simplereasoning}. 

In order to instill strong reasoning and meta-reasoning capacities in agentic systems, we need to view reasoning processes in terms of the cognitive capacity and abilities of LLMs. We need to take our intuitive reasoning ability and precisely formalize it using three interlocking principles: 
\begin{enumerate}[
    leftmargin=3em,
    rightmargin=1em,
    itemsep=2pt,
    topsep=1pt,
    label=\textbf{P\arabic*:}
]
    \item Associative vs.~Formal -- delegating what we can to formal code.
    \item Object vs.~Meta -- decomposing object-level tasks using meta-reasoning until each associative object-level task fits within the capacity of an LLM.
    \item Atomic vs.~Monolithic -- decomposing the meta-reasoning, which is itself cognitively demanding, into composable atomic units that can be delegated to different context threads.
\end{enumerate}

\subsection{Formal Notation}
\subsecvs
In order to precisely formalize reasoning ability along the three dimensions (\textbf{P1}, \textbf{P2}, \textbf{P3}) mentioned in \S\ref{sec:design-principles}, we now introduce the formal language of Deep Reasoning. 

\begin{table}[t]
\centering
\small
\renewcommand{\arraystretch}{1.2}
\begin{tabular}{l p{3.6cm} l}
\toprule
\textbf{Symbol} & \textbf{Description} & \textbf{Example} \\
\midrule

$\Ical$ & True Interpreter
& $\Ifn{x = `an'\bsep\ `S\{x\} Fr\{x\}cisco'} = $ \\
& & \indii $\Ifn{The City by the Bay} = \text{\openquote San Francisco\closequote{}}$ \\

$\Acal$ & Associative Interpreter
& $\Afn{ The City by the Bay} = \text{\openquote San Francisco\closequote{}}$ \\

$\Ecal$ & Formal Interpreter
& $\Efn{}{ x = `an'\bsep\ `S\{x\} Fr\{x\}cisco'} =$ \text{\openquote San Francisco\closequote{}}  \\

% $\MFmod{}{x}$ & Mental models of $x$ 
% & $\text{\openquote The City by the Bay\closequote{}} \in \Mmod{\text{San Francisco}}$ \\

$\MFmod{\Ecal}{x}$ & Formal models of $x$ w.r.t. \Ecal
& \text{\openquote x = `an'\bsep\ `S\{x\} Fr\{x\}cisco'{\closequote{}}}$\, \in \MFmod{\Ecal}{\text{\openquote San Francisco\closequote{}}}$ \\

$\mhl$ & Formal modeling function 
& $\mhl(\text{\openquote sum of 3 and 4\closequote{}}) = \text{\openquote add(3,4)\closequote{}}$ \\

$\IDeep$ & Deep Reasoning
& $\IDo \text{\openquote The zip code of the City by the Bay\closequote{}} \IDc =$ \\
% & $\IDeep \textcolor{RawSienna}{\big (} \text{\openquote The zip code of the City by the Bay\closequote{}} \IDc =$ \\
& & $ 
\hspace{5pt}\Efn{}{y=\Afn{City by the Bay}\bsep \hspace{2pt}
\IDeep\textcolor{RawSienna}{(}{\openquote zip code of \{y\}\closequote{}}\textcolor{RawSienna}{)}}$ \\
\bottomrule \\
\end{tabular}
\caption{High-level summary of the language of Deep Reasoning.}
\label{tab:notation}
\end{table}

\parvs
\paragraph{Symbols and Interpretations.}
We begin with a few basic definitions. A \emph{symbol} is a word such as ``hot'' or ``dog'', or a mark such as ``$+$'' or ``\&'', that stands for some abstract concept. Given a set of symbols $\mathbf{\Sigma}$, a \emph{sentence} $x$ over $\mathbf{\Sigma}$ is a sequence of symbols $x \in \mathbf{\Sigma}^+$. 

An \emph{interpretation} function $f$ is a function that maps sentences to concepts, i.e., $f:\bm{\Sigma}^+\to \bm{C}$, where $\bm{C}$ is the set of all concepts. Concepts can be represented in many different ways including sentences, images, and more. For simplicity, we limit our discussion to concepts represented by sentences.
We denote by $\Ical$ the ideal \emph{true interpretation}, which maps each sentence to its correct concept. In practice, we do not have access to $\Ical$. For example:
\begin{align*}
    \Ifn{The City by the Bay} & =\Ifn{SF, CA} = \Ifn{San Francisco} \;=\; \text{\openquote San Francisco\closequote{}}
\end{align*}
\parvs
\paragraph{Formal \& Associative Interpreters (P1).} In the language of Deep Reasoning, the distinction between associative and formal reasoning is explicit. Let $\mathbf{L}\subset \mathbf{\Sigma}^+$ be some formal language and $\Ecal:\mathbf{L}\to \mathbf{\Sigma}^+$ be a \emph{formal interpreter} that can execute sentences in $\mathbf{L}$. For example, \mbox{$\Efn{}{x=1; x+2}=\openquote{} 3\closequote{}$}. We extend $\Ecal$ with external functions that can be invoked during execution:
\begin{align*}
\Efn{\text{add}}{add(3,4)} = \openquote{}7\closequote{} \quad \quad \Efn{\text{llm}}{llm('City by the Bay')}=\openquote{} \text{San Francisco}\closequote{}
\end{align*}
On the other hand, we say that $\Acal$ is an \emph{associative interpreter} if it uses associative reasoning to assign meaning to sentences. Examples of associative interpreters include LLMs and other neural networks. Unlike $\Ecal$, $\Acal$ does not require formal sentences but also does not guarantee correctness or internal consistency. For example, ``San Francisco'' can be modeled both associatively and formally:
\begin{align*}
    \Afn{The City by the Bay} = \Efn{}{x = `an'\bsep \, `S\{x\} Fr\{x\}cisco'} = \text{\openquote San Francisco\closequote{}}
\end{align*}
\parvs
\paragraph{Formal Models \& Modeling Functions (P2).} When humans meta-reason about a task, we create a step-by-step decomposition on how to solve it (Figure~\ref{fig:conceptual_onion}). Therefore, the language of Deep Reasoning is able to capture such meta-reasoning processes. To this end, we define the set of all \emph{formal models}\footnote{We distinguish mental models from Machine Learning (ML) models, and call the former \emph{models} throughout this paper.} of $x$ under $\Ecal$ as all formal sentences that, when executed, map to the same underlying concept as $x$:
% We denote the set of all formal models of $x$ under $\Ecal$ as: 
% \pw{don't leave spaces before and after align blocks like this, it introduces extraneous white space}
%
\begin{align*}
\MFmod{\Ecal}{x} = \bigl\{ y \in \bm{L} \;\big|\; \Eo y \Ec = \Io x \Ic \bigr\} 
\end{align*}
Given a formal model of $x$ denoted as $y\in \MFmod{\Ecal}{x}$, we call $(x,y)$ a \textit{formal decomposition} of $x$ into $y$. To make this more concrete, in our running example, the converted human meta-reasoning trace (Figure~\ref{fig:conceptual_onion}.c) is a formal decomposition of the task ``Which volleyball court in...''.

%
% \bn{When a task is referred to colloquially as a \emph{formal task}, it means that it is trivial...\sout{When one refers to a task described by a sentence $x$, colloquially, as a \emph{formal task}, they mean that it is \bn{\sout{trivially easy} trivial}}} 
When a task is referred to colloquially as a \emph{formal task}, it means that it is trivial to translate the task-sentence into a formal model. For example, given $x = $ ``Which set of numbers in $\{\{6,1,2\},\{6,7\}\}$ has the most elements over 5?'', we can easily model it formally as:
\begin{align*}
y \, = \, \text{\openquote }\mathcal{X} = \{\{6,1,2\},\{6,7\}\}\bsep\ 
\operatorname*{argmax}_{S \in \mathcal{X}} 
|\{ x \in S \mid x > 5 \}|
\text{\closequote{}} 
\end{align*}
%
% Note that $(x,y)$ is a \textit{formal decomposition} since both $y$ and $x$ model the same concept: 
% \[ y \in \MFmod{\Ecal}{\openquote{}\{6,7\}\closequote{}} \]
%\[ x \in \MFmod{}{\openquote{}\{6,7\}\closequote{}} \quad \quad y \in \MFmod{\Ecal}{\openquote{}\{6,7\}\closequote{}} \]
We define a \textit{modeling function} as $\mhl: \bm{\Sigma}^+ \to \bm{L}$ such that for every $x\in \bm{\Sigma}^+$, $\mhl(x) \in \MFmod{\Ecal}{x}$. In other words, a modeling function takes a sentence and formalizes it into an \emph{equivalent} sentence that can be executed. Modeling functions can be extended with external functions:
\begin{align*}
\mfn{\text{add}}{the sum of 3 and 4} =\text{\openquote add(3,4)\closequote{}}
\end{align*}
Here, the formal model ``add(3,4)'' is interpretable by $\Ecal_{\text{add}}$. We can consider more abstract functions, such as the associative function $\Acal$. In such cases, the modeling function $\mhl_{\Acal}$ can formalize sentences using associative functions. For example,
\begin{align*}
\mfn{\Acal}{Number of a's in the City by the Bay}
= \openquote{}\text{\Afn{The City by the Bay}.count(`a')}
\closequote{}
\end{align*}
In this case, our modeling function $\mhl_{\Acal}$ decided to delegate the disambiguation of ``The City by the Bay'' to associative reasoning via $\Acal$, after which the $\texttt{count}$ function counts the number of a's.

\parvs
\paragraph{Incremental Modeling (P3).} Up until now all modeling examples assume an $\mhl$ capable enough to fully formalize a task in a single step. However, modeling in itself is a difficult cognitive task. Therefore we want to capture the idea of \emph{iteratively refining} a sentence until it becomes fully formal.
% \bn{This is great! I think all of these definitions should have little motivations like this.} 
% \bn{I'd avoid scare quotes in explanations---they are usually a mark of imprecise language. In this case, I don't think you need them - you can just say \textit{call} or \textit{invoke}} 

To do this, we allow the modeling function $\mhl$ to call itself within a formalization. This means it can produce intermediate formal sentences that still contain unresolved parts, which are then handled in subsequent modeling steps. To make this concrete, consider the example:
% A modeling function with a modeling oracle can capture this process. \pw{I'm confused by why we need to talk about a modeling function with a modeling oracle here.} To make this concrete, consider the example:
%
\begin{multline*}
\mfn{\mhl}{Number of tie-breaks in the following volleyball scoreboard: 3-1, ..., 2-3} = \\
\text{\openquote{}}\text{s = \{‘3-1', ..., ‘2-3'\}}\bsep \,\, \text{re = } \mfn{}{regex for volleyball tie-breaks} \bsep\,\, \text{len(s.filter(re.match))\closequote{}}
\end{multline*}
In this case, coming up with the regex for capturing ``volleyball tie-breaks'' might involve looking up the latest official volleyball rules from FIVB\footnote{FIVB (Fédération Internationale de Volleyball) is the governing body responsible for all forms of volleyball globally.}, building out several regex candidates and verifying them using a programming environment. This cannot be done reliably by a single associative function which is why $\mhl_{\mhl}$ delegated the task to a subsequent modeling step. 

\parvs
\paragraph{Summary.}
With the language of Deep Reasoning we can now express delegation between formal and associative tasks (\textbf{P1}, \S\ref{sec:design-principles}) and even formalize dependencies between associative subtasks using $\Acal$ and $\Ecal_{\Acal}$. We can express meta-reasoning using $\mhl_{\Acal}$ to decompose object-level tasks until the load delegated to an associative call is below the cognitive load of a given LLM context thread (\textbf{P2}, \S\ref{sec:design-principles}). Lastly, we can decompose the meta-reasoning process itself, using $\mhl_{\mhl}$, into nested layers of formal and associative meta-reasoning. This separates the high cognitive load of meta-reasoning between different LLM context threads (\textbf{P3}, \S\ref{sec:design-principles}).
% \pw{At this point, the reader just read through 2 pages of definitions and the section abruptly ends! It then transitions to introducing the principles of deep reasoning which superficially sounds like it could have been done without any of these definitions.}

\section{\DR{} -- A Deep Reasoning Agent}\label{sec:deep_reasoning}

\secvs
We can define specific Deep Reasoning architectures in terms of the high-level meta-reasoning using Deep Reasoning, separating the architecture from specific implementation decisions.
In this section we present a specific Deep Reasoning agent we call \DRb{} (Deep Layered Reasoning Scaffold).

%\bn{\sout{the}}
\subsection{Architecture} \label{sec:arch}

\subsecvs
We recursively define \DR{} ($\IDeep$) and an \textit{atomic modeling} function $\mhl^{a}$ as follows:
\begin{gather}
    \text{\IDo} x \text{\IDc} = \Ecal_{\IDeep,\,\Acal}\!\textcolor{blue}{\bigl(}\mhl^a_{\IDeep,\,\Acal}\textcolor{OliveGreen}{(}x\textcolor{OliveGreen}{)}\textcolor{blue}{\bigr)}
    \label{eq:deep_reasoning} \\[0.4em]
    \mhl^a_{\IDeep,\Acal}(x) = \text{\Ao\openquote{}decompose $\{x\}$ into a small formal model over } \notag \\
    \text{ $\IDeep,\Acal$ using the following atomic decompositions \ldots \closequote{}\Ac} \notag
\end{gather}
%
% oracles \pw{`oracle' implies something that we don't actually have in practice so I'm confused by why this is appearing instead the implementation section. The notion of it being an oracle doesn't seem necessary?} of 
In a nutshell the \DR{} $\IDeep$ takes as input a sentence $x$ and \emph{recursively} decomposes it and executes it through an atomic modeling function. Notably, the modeling function $\mhl^{a}$ is equipped with both the associative function $\Acal$ and the Deep Reasoning agent $\IDeep$ itself. 

At each step, $\IDeep$ takes a sentence, decomposes it into an atomic formal model using $\mhl^a_{\IDeep,\Acal}$, and executes it via $\Ecal_{\IDeep,\Acal}$, recursively invoking itself on sub-calls as needed. In-context atomic decompositions are fed into the system prompt of $\Acal$, allowing fine-grained control of the decomposition patterns that the agent will see without additional programming.
This leads to a nested cascade of formal, associative and meta-reasoning steps (see Figure~\ref{fig:conceptual_onion}) that can be controlled by the user. 
For brevity, in the following sections, we shorten $\Ecal_{\IDeep,\,\Acal}$ to $\Ecal$ and $\mhl^a_{\IDeep,\Acal}$ to $\mhl$.

% \dl{try to change onion macros to textSC}
% If $\mhl(x)=y$ for some sentence $x$ and a formal model $y$, we denote the interpreting of $x$ through execution of $y$ as:
% %
% \begin{align*}
%     \text{\IDo} x \;\text{\IDca}\; \text{\Eo} y \text{\Ec} \text{\IDc}
% \end{align*}
% %

\subsection{Implementation}\label{sec:implementation}

\subsecvs
In this section, we describe the implementation details of the \DR{} agent $\IDeep$, and explain how to build general-purpose agents using the language of Deep Reasoning.

%\bn{\sout{have to }}, \bn{\sout{and we do it }}
\parvs
\paragraph{Instantiation.} So far, the modeling function $\mhl$, associative function $\Acal$, and formal interpreter $\Ecal$ have been conceptual. To operationalize Equation~\eqref{eq:deep_reasoning} in the \DR{} agent, we instantiate them as follows:
\begin{center}
$\Ecal$ is Python \quad
$\Acal$ is an LLM \quad
$\mhl$ is an LLM
\end{center}
Importantly, as specified in Equation~\eqref{eq:deep_reasoning}, the formal interpreter $\Ecal$ (Python) has access to $\Acal$ (LLM) and the \DR{} ($\IDeep$) itself as external functions. This allows the agent to recursively call itself inside the environment (Figure~\ref{fig:conceptual_onion}). The Python environment can also include additional tools and variables.

The modeling function $\mhl$, implemented as an LLM, is responsible for decomposing tasks into a series of associative steps (calls to $\Acal$), formal steps (code executed by $\Ecal$), and meta-reasoning steps (recursive calls to $\IDeep$). To guide this behavior, we design prompts for the modeling LLM ($\mhl$) with \textit{atomic} decompositions as in-context examples. These atomic decompositions are built by first formalizing human meta-reasoning traces to the language of Deep Reasoning and then converting them into in-context examples.

\parvs
\paragraph{Human Meta-reasoning Traces.} We start from verbalized human meta-reasoning traces for specific tasks. For example, in the volleyball court task ``Which volleyball court in ...'', humans have an intuitive ability to meta-reason and decompose the task, as illustrated in Figure~\ref{fig:conceptual_onion}. These traces are directly mapped in the language of Deep Reasoning (Figure~\ref{fig:conceptual_onion}.c), where different parts of reasoning map to different interpretation functions. For instance, resolving ``City by the Bay'' to San Francisco is handled associatively via $\Acal$. The step of ``I will list all the volleyball courts in SF'' is expressed as a recursive call to $\IDeep$, and ``pick the court with the most tie-breaks'' is a simple formal operation (\texttt{argmax}). Basically, this formalized human meta-reasoning \emph{defines} how the task should be decomposed by the modeling LLM ($\mhl$).

\parvs
\paragraph{In-Context Examples.} Given these formalized traces, we convert them directly into in-context examples. Associative calls $\Acal$ and Deep Reasoning calls $\IDeep$ are implemented as LLM calls and recursive sub-agent calls, respectively. Figure~\ref{fig:conceptual_onion}.d shows this concretely: the code corresponds to a single in-context example that specifies how to decompose the ``Which volleyball court in ...'' task based on the formalized human meta-reasoning trace (see \S\ref{apn:prompts} for details). This example is then provided to the modeling LLM ($\mhl$). Generally, in-context examples derived from human meta-reasoning on different tasks (e.g., long-context reasoning, multi-hop QA) can be added in the same way, allowing the modeling LLM ($\mhl$) to generalize decomposition strategies across tasks.

\parvs
\paragraph{Agentic Loop.} Finally, we implement the agent inside a Python Read-Eval-Print Loop (REPL), similar to a notebook environment. Each step consists of a short chain-of-thought from $\mhl$, followed by a formal code block that is executed. The REPL allows the agent to store, inspect, and reuse variables across steps. Recursive calls to \DR{} ($\IDeep$) run in separate REPL environments.

\section{Experiments}\label{sec:experiments}

\begin{table}
\centering
\hspace*{-0.65cm}
\begin{tabular}{l l cccc c}
\cmidrule(lr){2-7}
\multirow{3}{*}{} & \multirow{3}{*}{\diagbox[width=11em]{\textbf{Benchmark}}{\textbf{Method}}}
& \multirow[c]{3}{*}{\raisebox{0.7em}{\textbf{ReAct}}}
& \multirow[c]{3}{*}{\raisebox{0.7em}{\textbf{CodeAct}}}
& \textbf{Deep}
& \multirow[c]{3}{*}{\raisebox{0.7em}{\textbf{RLM}}}
& \multirow[c]{3}{*}{\raisebox{0.7em}{\textbf{\DR{}}}} \\
& & & &\textbf{Research} &  & \\
& 
& \raisebox{0.35em}{\citeyearpar{yao2022react}}
& \raisebox{0.35em}{\citeyearpar{wang2024executable}}
& \raisebox{0.12em}{\citeyearpar{roucher_open_deep_research_2025}}
& \raisebox{0.35em}{\citeyearpar{zhang2025recursive}}
& \raisebox{0.35em}{\textit{(ours)}} \\
\cmidrule(lr){2-7}
\multirow{4}{*}{\rotatebox{90}{\small\textit{\shortstack{Qwen3 8B \\ Thinking}}}}

& SynthWorlds~\citeyearpar{gu2025synthworlds}
   & 0.176 & 0.268 & 0.206 & 0.058 & \textbf{0.305} \\
& PhantomWiki \citeyearpar{gong2025phantomwiki}
   & 0.144 & 0.153 & 0.120 & 0.043 & \textbf{0.172} \\
& DeepSearchQA \citeyearpar{gupta2026deepsearchqa}
   & 0.124 & 0.149 & 0.142 & 0.139 & \textbf{0.161} \\
& Oolong-\textit{real}~\citeyearpar{bertsch2025oolong}
   & NA$^\dagger$ & 0.045 &  NA$^\dagger$ & 0.065 & \textbf{0.076} \\
\cmidrule(lr){2-7}
\multirow{4}{*}{\rotatebox{90}{\small\textit{\shortstack{Qwen3 32B \\ Thinking}}}}

& SynthWorlds~\citeyearpar{gu2025synthworlds}
   & 0.228 & 0.281 & 0.275 & 0.169 & \textbf{0.346} \\
& PhantomWiki \citeyearpar{gong2025phantomwiki}
   & 0.167 & 0.252 & 0.212 & 0.160 & \textbf{0.369} \\
& DeepSearchQA \citeyearpar{gupta2026deepsearchqa}
   & 0.177 & 0.213 & 0.230 & 0.190 & \textbf{0.241} \\
& Oolong-\textit{real}~\citeyearpar{bertsch2025oolong}
   & NA$^\dagger$ & 0.060 & NA$^\dagger$ & 0.077  & \textbf{0.132} \\
\cmidrule(lr){2-7}
\multirow{4}{*}{\rotatebox{90}{\small\textit{\shortstack{Llama-3.3 \\ 70B Instruct}}}}

& SynthWorlds~\citeyearpar{gu2025synthworlds}
   & 0.300 & \textbf{0.480} & 0.308 & 0.401 & 0.359 \\
& PhantomWiki \citeyearpar{gong2025phantomwiki}
   & 0.244 & 0.381 & 0.160 & 0.270 & \textbf{0.512} \\
& DeepSearchQA~\citeyearpar{gupta2026deepsearchqa}
   & 0.155 & 0.184 & 0.127 & 0.175 & \textbf{0.187} \\
& Oolong-\textit{real}~\citeyearpar{bertsch2025oolong}
   & NA$^\dagger$ & 0.079 & NA$^\dagger$ & 0.064 &  \textbf{0.151} \\
\cmidrule(lr){2-7} \\
\end{tabular}
\vspace{0.5mm}
\caption{Scores (0--1, $\uparrow$ higher is better) across four reasoning benchmarks. NA$^\dagger$ indicates context size limitations. \DR{} outperforms state-of-the-art methods across all benchmarks by an average of 24.8\% over the best-performing baseline.}
% \mt{remove surrounding lines}
\label{tab:main-results}
% \vspace{-5mm}
\end{table}

\subsection{Setup}

\subsecvs
To empirically evaluate the utility of Deep Reasoning in agentic reasoning tasks, we take four difficult reasoning benchmarks with verifiable answer that test different reasoning types. 

\parvs
\paragraph{Benchmarks.} The benchmarks include: \circled{1} \textbf{SynthWorlds} \citep{gu2025synthworlds} tests multi-hop reasoning over synthetic knowledge-graph–derived documents. This benchmark can isolate reasoning from memorization, requiring models to not rely solely on parametric knowledge and test if they ground on external knowledge reliably. \circled{2} \textbf{PhantomWiki} \cite{gong2025phantomwiki} tests multi-hop QA over a synthetic universe of configurable size. This benchmark tests how models track intermediate states across long reasoning chains.  \circled{3} \textbf{DeepSearchQA} \citep{gupta2026deepsearchqa} is a verifiable deep research benchmark. It has complex, multi-step information-seeking tasks that test an agent's ability to gather, filter and organize information.  \circled{4} \textbf{OOlong-(real)} \cite{bertsch2025oolong}  tests multi-step information aggregation and state tracking over very long real-world documents (e.g., episode transcripts). This benchmark tests how agents deal with documents that go beyond a base model's context window size.
% \kg{, and requires maintaining state (counts, indices, and aggregates) across multi-episode transcripts spanning up to millions of tokens}. 
% \bn{Just noting: reviewers might ask for RLM with coding models like Qwen3-Coder-Next, or for comparisons using api models.} \dl{We will run these between submissions and rebuttals, we didnt have the compute before}

\parvs
\paragraph{Baselines.} We compare our method against four open source baselines: \circled{1} A ReAct \citep{yao2022react} implementation from~\cite{smolagents}. \circled{2} A {CodeAct~\citep{wang2024executable}} implementation from~\cite{smolagents}. \circled{3} A Deep Research agent from \citep{roucher_open_deep_research_2025}. \circled{4} The RLM agent from the paper \cite{zhang2025recursive}.

\parvs
\paragraph{Models.} We test the baselines and our proposed method on the benchmarks using 3 different models from 2 model families: Qwen3-8B Thinking, and Qwen3-32B Thinking~\citep{yang2025qwen3}, Llama-3.3 70B Instruct~\citep{dubey2024llama}. See \S\ref{apn:exp_details} for more details.

\parvs
\paragraph{Constructing decompositions} To create decompositions for each benchmark, we followed the process outlined in Figure \ref{fig:conceptual_onion}. We took 2-3 made up tasks representative of the tasks in each benchmark and verbalized our own reasoning about how to solve them. We then translated the decompositions to the language of deep reasoning while making sure they are as atomic as possible. This lead to 5-10 decomposition examples per benchmark which we then wrote in \DR{}'s syntax.

\subsection{Results and Analyses}

\subsecvs
\paragraph{\DR{} outperforms baselines across different benchmarks.}
As shown in Table~\ref{tab:main-results}, \DR{} outperforms all evaluated scaffolds in $11/12$ settings, with average improvements of 36.4\% (Qwen3-32B), 12.8\% (Qwen3-8B), and 25.4\% (Llama-3.3-70B) over the strongest baseline on each task. An exception is CodeAct outperforming \DR{} on SynthWorlds with Llama-3.3-70B. Notably, \DR{} consistently punches above its model class, with the 8B model outperforming the best evaluated 32B baseline on SynthWorlds and Oolong-real, and the 32B model outperforming the best evaluated 70B baseline on DeepSearchQA and OOlong-real.
% As can be seen in table \ref{tab:main-results}, \DR{} outperforms all other scaffolds by substantial in $11/12$ cases with an average improvement of score of 36.4\% (Qwen3-32B), 12.8\% (Qwen3-8B) and 25.4\% (Llama-3.3-70B) when comparing against the best performing baseline on each task.
% A notable exception is CodeAct outperforming \DR{} on the SynthWorlds benchmark with model Llama-3.3 70B. Notably, \DR{} clearly punches above its model class against the best performing model in synthworld (8B vs 32B) Oolong-real (8B vs 32B) and PhantomWiki (32B vs 70B).

\parvs
\paragraph{Reducing cognitive load mitigates premature termination and hallucination.}
To better understand the gains of \DR{}, we perform a multi-label topic-based failure analysis on Qwen3-32B traces from \DR{} and the baselines (see \S\ref{apn:failure_mode} for details). Despite the unlimited toekn budget given to all baselines, the two dominant failure modes are \textbf{premature termination} (78\%) and \textbf{hallucination} (45\%). For premature termination, we observe ReAct and CodeAct attempting to resolve answers in one or two hops on SynthWorlds and PhantomWiki, often giving up when these high-cognitive-load reasoning steps fail. For hallucination, we observe RLMs delegating large counting tasks to sub-LLMs that count unreliably, while both RLMs and Deep Research fabricate mock tool outputs and entities in SynthWorlds and PhantomWiki. Manual inspection of traces shows that \DR{} avoids these pitfalls by decomposing the same tasks in more atomic ways, leading to correct solutions in cases where other scaffolds fail. We provide qualitative trace comparisons in \S\ref{apn:qualitative_examples} and recursion depth analysis in \S{\ref{app:recursion-depth}}. Analyzing both total and per-thread token counts reveals a clear pattern. 

\parvs
\paragraph{\DR{} makes effective use of available inference-time compute.}
As expected, \DR{} spawns many reasoning threads, resulting in 12.9$\times$ more tokens on average. This is despite the fact the all methods were run with infinite budget compute. However, on a per-thread basis, \DR{} reduces reasoning tokens by an average of \textbf{71\%} and non-reasoning tokens by \textbf{76\%}. These results are consistent with our hypothesis that reducing cognitive load mitigates premature termination and hallucination. To verify that the gains of \DR{} do not come simply from the usage of more tokens, we re-run the strongest baseline on each benchmark with Qwen3-32B in a token-matched regime. In this regime, we do majority voting over $K$ sampled answers from the baseline, where $K$ is the ratio of \DR{}'s average token usage to the baseline's (rounded up). Majority voting changes baseline scores by at most 0.024, and \DR{} outperforms the token-matched baseline on all four benchmarks (see \S\ref{apn:token-matched} for details). This indicates that \DR{}'s gains stem from how compute is distributed across reasoning threads rather than from the total amount of compute. While these gains currently come at a high token cost, we hypothesize that \circled{1} the high overlap of system prompts across many LLM threads, \circled{2} the ability of \DR{} to bridge scaling tiers by using cheaper models, and \circled{3}~selecting models based on the cognitive load of associative tasks, could eventually lead to lower parameter-adjusted token costs.

\parvs
\paragraph{Deep Reasoning performance is enabled by in-context decomposition examples.} 
% \mt{can we have a more understandable/catchy title?  ``Deep Reasoning is what drives improvements'', or sth like that.}\dl{All of this system is Deep Reasoning in a sense, we can maybe change it to, "in-context decomposition drive performance in Deep Reasoning"  } 
A natural question is whether \DR{}' improvements come from the structured meta-reasoning in-context examples themselves, or simply from good engineering. A follow-up question is whether describing in NL of how to \textit{atomically} decompose is sufficient for current LLMs to operationalize it effectively. We ablate this on Qwen3-32B by running two reduced variants. The first, \textit{no-examples}, removes all decomposition examples from the system prompt. The second, \textit{with-principles}, also removes the decomposition examples, but replaces them with NL instructions for decomposing tasks along the three dimensions listed in \S\ref{sec:design-principles}.
The full ablation table is in Appendix Table~\ref{tab:ablation-examples}. We see that removing the in-context examples causes large drops on every benchmark (70\% on average). Interestingly, \textit{with-principles} performs even worse than \textit{no-examples} in F1 on every benchmark (0.117 vs. 0.139 on DeepSearchQA; 0.080 vs. 0.113 on PhantomWiki; 0.036 vs. 0.041 on SynthWorlds; 0.033 vs. 0.036 on Oolong). These results suggest that current LLMs are not good at operationalizing structured meta-reasoning and require humans to meta-reason about task decompositions. LLMs appear to be reducing the task of structurally decomposing a task to pattern matching. We discuss the need for high quality decompositions, the expertise required to create good decompositions and other limitations and future work directions in \S\ref{sec:limitations}.

\section{Related work} \label{sec:related_work}

\parvs
\paragraph{Train Time Scaffold Generation.}
Concurrent work~\citep{metaharness2026,recreate2026} build novel scaffolds automatically by using an outer training loop that uses LLMs to search over the space of scaffolds during training, outputting scaffolds that outperform manually curated scaffolds on the tasks for which they were trained. Train-time scaffold generation is a promising direction, but must contend with data collection and continual retraining costs as out-of-distribution tasks come up. \DR{} tries to avoid this cost by in-context learning from a small number of atomic decompositions. 

\parvs
\paragraph{Test time Scaffolds Generation.}
Continual-learning scaffold generators adapt over time by performing additional inference-time LLM calls to reflect, summarize, and store experience between sessions, using this experience to improve subsequent scaffold generations per task~\citep{masonthefly26,liveswe25}. \cite{masonthefly26} learns to decompose tasks into different \textit{multi agent interaction patterns} which are then supervised and adapted by a watcher agent which interacts with an experience pool. \citep{liveswe25} focus on coding tasks, starting with a mini-SWE-agent that only has bash tools, and continually expands the set of tools the agent has through experience, without modifying the agentic loop.
Continual learning of decompositions is a promising direction complementary to our work. While both works learn from examples well, it is unclear how to inject human expertise into this process. 
\DR{} directly translates human reasoning using the Deep Reasoning language, which makes it easily controllable via prompting with a small set of in-context examples. Moreover, instead of dynamically varying tools and workflow patterns between tasks, \DR{} also varies the scaffold during task execution, adapting to changes in the intermediate state of the task progression.

\parvs
\paragraph{Recursive and Programmatic Decomposition.}
Several recent methods also structure reasoning through code or recursion, but they differ in which part of the structure is fixed in advance. REPL-Plan~\citep{liu2024interactiveexpressivecodeaugmentedplanning} organizes planning as parent and child LLM-REPLs with generic planning instructions, a setting close to our \textit{no-examples} ablation (\S\ref{apn:ablation-examples}). Atom of Thoughts~\citep{teng2025atom} applies a fixed meta-reasoning policy of decompose-and-contract to a question at every step through associative LLM calls, with no formal execution for quantitative subproblems. Concurrent work on $\lambda$-RLM~\citep{roy2026mathbfycombinatorllmssolvinglongcontext} constrains RLMs~\citep{zhang2025recursive} to a fixed set of decomposition operators shared across tasks. In contrast, \DR{} constructs the decomposition at test time from in-context examples and adapts it to the task and to intermediate results.

\section{Conclusions}\label{sec:conclusions}

% As humans, we are able to solve tasks by planning, executing, revising intermediate goals, resolving ambiguity via associative links and applying formal procedures. This flexibilty is lacking in modern agentic systems: they have been hardcoded to specific patterns of reasoning. As a result, they struggle when tasks require adapting the structure of reasoning. In this work, we introduce the Deep Reasoning paradigm, a way to capture human meta-reasoning traces and build ``just-in-time'' meta-reasoning scaffolds. We also introduce the language of Deep Reasoning, a formal way to capture human meta-reasining traces that can be directly translated as in-context examples for agents. We operationalize our paradigm in a Deep Reasining agent, \DR, and showcase improvements over all evaluated baselines. 

% \bn{Emphasizing that other scaffolds are ``manually curated'' might raise the objection that our in-context examples also manually curated. Maybe saying something like ``purpose-built task-specific scaffolds''.}\dl{I change to manually constructed scaffolds, does it work now?}.
Humans solve complex tasks by planning, executing, revising intermediate goals, resolving ambiguity and switching between associative and formal reasoning.
This flexible just-in-time meta-reasoning is what current LLM scaffolds lack, leaving them brittle against novel tasks.
This work operationalizes human meta-reasoning in general purpose agents through Deep Reasoning, a formal language for structured meta-reasoning.
Through Deep Reasoning we construct \DR{}, an agentic scaffold that evolves just-in-time based on the specific task and intermediate reasoning steps.
Using a few atomic in-context examples, based on human intuition, \DR{} significantly outperforms strong baseline scaffolds on a collection of hard reasoning tasks.  Under the Deep Reasoning paradigm, LLMs should not be viewed as a collection of shallow experts, to be managed by scaffolding glue, but associative reasoning processes that flexibly interact with formal processes to create artificial minds that can meta-reason robustly. Even a basic implementation of Deep Reasoning can bridge the scaling gap (8B vs 32B).
% \mt{sounds kinda defensive?  maybe "Even a \textit{basic} implementation of Deep Reasoning can bridge the scaling gap (8B vs 32B)." and let readers infer on their own how interesting this is.}

\paragraph{Limitations and Future Work.}\label{sec:limitations}
\DR{}'s performance gains are predicated on having high quality Deep Reasoning decompositions, and it cannot generalize in zero shot to novel domains that require substantially different decompositions. Creating such decompositions requires good understanding of the Deep Reasoning language, the abilities of modern LLMs and agentic loop structures. This might present a barrier of entry for experts in other domains for writing decompositions for their use cases. Additionally, complex tasks in domains like scientific or medical reasoning might require a lot of atomic decompositions, which could overwhelm a single memory thread. Future work can include methods for \circled{1} handling large collections of decompositions, \circled{2} reducing the token count, cost, and latency of Deep Reasoning agents through techniques such as caching shared prompts and routing lower-cognitive-load subtasks to smaller models, \circled{3} sourcing decompositions automatically from natural language feedback and textbooks, \circled{4} combining Deep Reasoning with continual learning from examples, and \circled{5} extending the language of Deep Reasoning to capture subjective or ambiguous sentences.

\paragraph{Broader Impact.}\label{sec:impact}
\DR{} has the potential to lower the barrier for domain experts to build reliable reasoning agents in fields such as science, medicine, and law by encoding expertise as a handful of in-context decompositions rather than as training data or predefined scaffolds. However, it still inherits the standard risks of LLM agents, as they are a component of the system. This can still include hallucination and confidently wrong intermediate steps, and its higher token cost concentrates capability in actors with large compute budgets. We mitigate these risks by evaluating only on public benchmarks, releasing all decomposition examples and traces, and recommending human-in-the-loop deployment for high-stakes use.

% MoE Deep Learning architectures overcome the representational capacity of single neural network layers by creating deep stacks on neural networks and varying their activations during inference. Similarly, Deep Reasoning proposes to overcome the cognitive capacity limitations of LLMs by creating deep nested layers that interleave many different memory threads with code in flexible ways during inference. 

% \mt{TODO: Check \textbf{citations}: \circled{1} conference verions (not arxiv), \circled{2} order of multiple citations (yulia said ``by relevance'', \circled{3} duplicates}

% \mt{Check surrounding fills around in the manuscript (use package?)}

% \mt{make sure we reference all appendices we have somewhere in main text.}

% \kg{Appendix C is empty} \mt{you didnt see my tiny little sentence :( }

% End of the main text.
\DRendmainmatter

\ifDRacks
\begin{ack}

% Michael was supported by NSF III 2507117 and NSF IIS 2314527.

This research was developed with funding from the Defense Advanced Research Projects Agency's (DARPA) SciFy program (Agreement No. HR00112520300). This material is based upon work supported in part by the Defense Advanced Research Projects Agency and the Air Force Research Laboratory, contract number(s): FA8650-23-C-7316. The views expressed are those of the author and do not reflect the official policy or position of the Department of Defense or the U.S.~Government. This research was also supported by the Coefficient Giving, Meta AIM program. This research was also supported by NSF III 2507117, NSF IIS 2554007 and NSF IIS 2314527. This work was also supported in part by the U.S. National Science Foundation (NSF) CAREER Award 2337877, Schmidt Sciences Award on AI \& Advanced Computing, through the Science of Trustworthy AI program, and by the University of Washington Tech Policy Lab. Any opinions, findings, and conclusions or recommendations expressed in this material are those of the authors and do not necessarily reflect those of NSF or Schmidt Sciences. This work was also supported in part by the Robert L. McDevitt, K.S.G., K.C.H.S. and Catherine H. McDevitt L.C.H.S. Chair in Computer Science at Georgetown University.
% Competing-interests half of the disclosure; required for the NeurIPS
% camera-ready, supplied by the driver (see main_neurips_final.tex).
\DRcompetinginterests
\end{ack}
\fi % \ifDRacks

\DRvenuestatements

\bibliography{main}

\appendix

\FloatBarrier

% \section{Test}

% Math Acal: $\Acal$, Math Afn: $\Afn{ The City by the Bay}$, No math Acal: \Acal, No math Afn: \Afn{x+1}

\section{Decompositions as Prompts}\label{apn:prompts}
\begin{figure}[t]
    \centering
\includegraphics[width=0.99\textwidth]{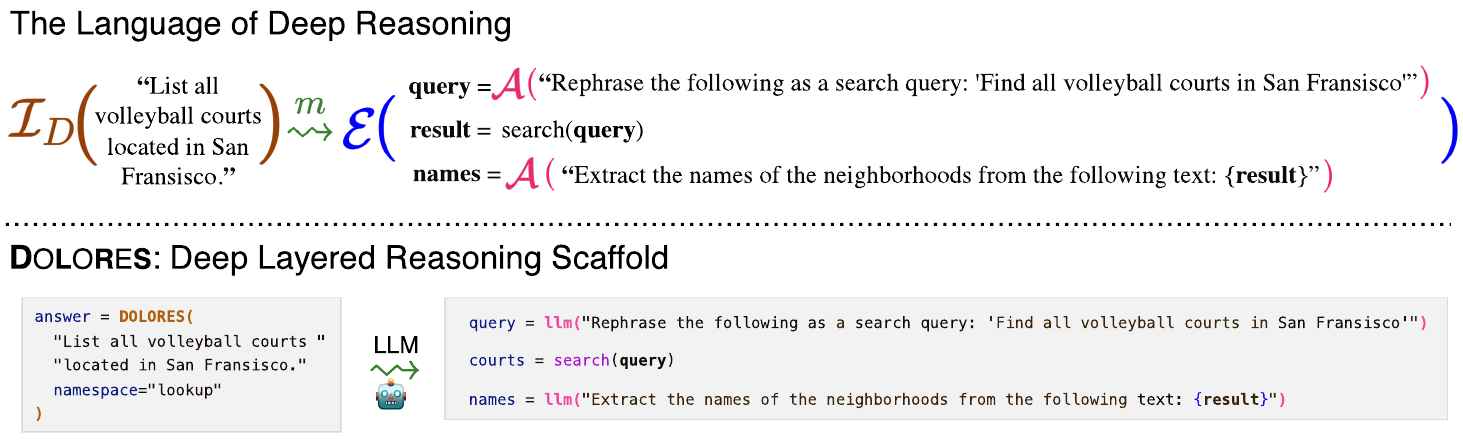}
    \caption{
    Low-level overview of Deep Reasoning on Running example. The original task is decomposed by $\mhl$ into sub-tasks, each resolved by a combination of associative reasoning ($\Acal$), formal execution ($\Ecal$), and recursive deep reasoning ($\IDeep$).}
    \label{fig:low-overview}
\end{figure}

In this we show the some of the decompositions generated from the running example in figures \ref{fig:conceptual_onion} and \ref{fig:low-overview}. 
Each few shot example shows a decomposition and comprises of a name-space (used to filter the set of in-context examples rendered to the system prompt of a \DR{} thread), a task and a series of "shallow reasoning" blocks, formal reasoning blocks and observation blocks. In order to keep the meta-reasoning repl blocks grounded on intermediate results, \DR{} inspects intermediate results as they execute to make sure that the model is executing as expected.
% ── sequential reasoning ──────────────────────────────────────────────────────
\begin{namespacebox}{sequential reasoning}
\task{Which volleyball court in the `City by the Bay' has hosted the most
tournament matches that went to tie-break?}

\vspace{4pt}
The nickname `City by the Bay' needs resolving first, then I'll enumerate courts,
look up each court's match history and count its tie-breaks, then pick the max.
\vspace{4pt}
\begin{repl}
city = llm("What city is known as the 'City by the Bay'? Return just the city name.")
print(city)
\end{repl}
\vspace{4pt}
\begin{observation}
San Francisco
\end{observation}
\vspace{4pt}
\begin{repl}
courts = dolores(
    f"List all volleyball courts located in {city}. Return a Python list of strings.",
    namespace="lookup"
)
print(courts)
\end{repl}
\vspace{4pt}
\begin{observation}
['East Beach Volleyball Courts', 'Crissy Field Beach', 'Marina Green Courts', 'South End Zone Courts']
\end{observation}

\vspace{4pt}
Now I fetch each court's match scores and count its tie-breaks.
\vspace{4pt}
\begin{repl}
tiebreaks = {}
for court in courts:
    scores = dolores(
        f"List all past tournament match scores for {court} in {city}."
        f" Return a list of match result strings (e.g. '3-2', '3-0').",
        namespace="lookup"
    )
    tiebreaks[court] = dolores(
        "How many of these volleyball match scores went to a tie-break?",
        scores=scores,
        namespace="formal"
    )
print(tiebreaks)
\end{repl}
\vspace{4pt}
\begin{observation}
{'East Beach Volleyball Courts': 14, 'Crissy Field Beach': 8, 'Marina Green Courts': 5, 'South End Zone Courts': 11}
\end{observation}
\vspace{4pt}
\begin{repl}
FinalAnswer(max(tiebreaks, key=tiebreaks.get))
\end{repl}
\end{namespacebox}

% ── lookup ────────────────────────────────────────────────────────────────────
\begin{namespacebox}{lookup}
\task{List all past tournament match scores for East Beach Volleyball Courts
in San Francisco. Return a list of match result strings (e.g.\ `3-2', `3-0').}

\vspace{4pt}
I'll paraphrase the task into a search query, retrieve results, then extract
structured scores with \texttt{llm}.
\vspace{4pt}
\begin{repl}
query = llm(
    "Rephrase as a concise web search query:\n"
    "Past tournament match results at East Beach Volleyball Courts San Francisco"
)
results = search(query)
print(results[:400])
\end{repl}
\vspace{4pt}
\begin{observation}
East Beach Volleyball Courts hosted the 2019 and 2022 CBVA Opens.
  2022 Open final:   Team A def. Team B  3-2
  2022 Open semi:    Team C def. Team D  3-0
  2022 Open quarter: Team E def. Team F  3-1
  2019 Open final:   Team G def. Team H  3-2  ...
\end{observation}

\vspace{4pt}
Results look good. Now extract the scores into a clean list.
\vspace{4pt}
\begin{repl}
scores = llm(
    "From the text below extract every match result as a Python list of strings"
    " in the format '3-0', '3-1', '3-2' (or '0-3', '1-3', '2-3' for losses)."
    " Return only the list.\n\n" + results
)
print(scores)
\end{repl}
\vspace{4pt}
\begin{observation}
['3-2', '3-0', '3-1', '3-2', '3-1', '2-3', '3-0', '3-2']
\end{observation}
\vspace{4pt}
\begin{repl}
FinalAnswer(scores)
\end{repl}
\end{namespacebox}

% ── formal ────────────────────────────────────────────────────────────────────
\begin{namespacebox}{formal}
\task{How many of these volleyball match scores went to a tie-break?
\vspace{4pt}
Variable \texttt{scores}: list of match result strings.}

\vspace{4pt}
I'll inspect the format first, then count matches with a 3-2 or 2-3 result ---
those are the only scorelines requiring a 5th tie-break set.
\vspace{4pt}
\begin{repl}
print(scores[:5])
\end{repl}
\vspace{4pt}
\begin{observation}
['3-2', '3-0', '3-1', '3-2', '2-3']
\end{observation}

\vspace{4pt}
Format confirmed. A tie-break is any match where 5 sets were played.
\vspace{4pt}
\begin{repl}
count = sum(1 for s in scores if s in ('3-2', '2-3'))
print(count)
\end{repl}
\vspace{4pt}
\begin{observation}
3
\end{observation}
\vspace{4pt}
\begin{repl}
FinalAnswer(3)
\end{repl}
\end{namespacebox}

\section{Baselines and Benchmarks Details} \label{apn:exp_details}

\subsection{Baselines}

\paragraph{ReAct~\citep{yao2022react} \& CodeAct~\citep{wang2024executable}.} We use the \texttt{ToolCallingAgent} and \texttt{CodeAgent} implementations from \texttt{smolagents}~\citep{smolagents}, respectively. These are production-oriented implementations of ReAct and CodeAct from \texttt{smolagents}~\citep{smolagents}, designed for real-world agentic systems rather than minimal reproductions of the original methods. In particular, they include surrounding engineering such as tool-call error handling, execution feedback, retry mechanisms, and corrective messages that help the LLM recover from malformed actions or invalid tool usage.

\paragraph{RLMs~\citep{zhang2025recursive}.} We use the official \texttt{rlms} implementation version \texttt{v0.1.1} (released 2026-02-18 on PyPI), which includes improvements released after the original paper submission.

\paragraph{Deep Research~\citep{openai_deep_research_2025}.} We compare against the open Deep Research implementation of~\cite{roucher_open_deep_research_2025}, which is also used as the Deep Research baseline in~\citep{lai2026kramabench}. The system consists of a manager agent coordinating web-search sub-agents equipped with practical browser tools for web search, document navigation, long-context inspection, archived page retrieval, and replanning. The agent operates over a shared browser state and supports multi-step information gathering across long web documents and files.

At the same time, compared to industrial Deep Research systems like NVIDIA's AIQ~\citep{nvidia_aiq_blueprint_2026}, this implementation remains relatively lightweight and transparent. In the literature, ``Deep Research'' systems are usually compared as full-stack implementations~\citep{shao2025dr}. That means they are not just comparing reasoning quality---they are comparing everything around it: orchestration design, shared workspaces, context management, source verification, URL sanitization, retry logic, routing, and all the engineering that makes these systems robust~\citep{openai_deep_research_2025, nvidia_aiq_blueprint_2026}. These scaffolds matter a lot. In practice, they drive a big part of the final performance. But if we take a closer look, the reasoning itself is often very similar across systems. Most follow the same pattern: a manager agent that spawns standalone search agents that go out, retrieve information, and come back. 

In this work, we propose a way to instill human meta-reasoning abilities in general purpose agents. Concretely, our contribution lies in the reasoning decompositions, not in the agentic scaffold surrounding it. \DR{} is a simple operationalization of Deep Reasoning. As a result, comparing directly against highly engineered Deep Research systems is misaligned, since it would largely measure the engineering behind those systems rather than the reasoning itself. Instead, we compare against ~\citep{roucher_open_deep_research_2025}. This provides a clean baseline: the same general decomposition used by top Deep Research systems (a manager agent that spawns specialized search sub-agents) but without the heavy engineering and often opaque optimizations. In other words, this setup isolates the comparison to the quality and power of the decompositions.

\subsection{Prompts}

We take particular care to provide strong prompts to the baselines on each benchmark. When possible, we use prompts that have already been calibrated in prior work. For PhantomWiki~\citep{gong2025phantomwiki}, we use the 10 in-context examples from the benchmark's ReAct experiments, and translate them to CodeAct's format (i.e., CodeAct expects actions enclosed in code XML tags instead of JSON tool calls) and to DeepResearch's format. RLMs instead use 17 REPL examples demonstrating how to use the provided tools. When calibrated prompts are not available, we construct them ourselves. SynthWorlds~\citep{gu2025synthworlds}, for example, does not provide an agentic baseline, but defines a taxonomy of six reasoning types (e.g., linear chain, diamond, etc.). We therefore handcraft ReAct examples covering all six types, translate them to CodeAct, and to DeepResearch; for RLMs, we instead provide examples demonstrating how to use the tools, as we do not want to impose specific decomposition patterns ourselves. On DeepSearchQA~\citep{gupta2026deepsearchqa}, we use the default prompts from~\citep{smolagents} for ReAct, CodeAct, and DeepResearch, and from the default implementation of RLMs~\citep{zhang2025recursive}. Finally, on Oolong-real, CodeAct and RLMs receive the default examples; we do not evaluate ReAct or DeepResearch on this benchmark due to context length overflow.

\subsection{Benchmark}

\paragraph{Oolong (real)~\citep{bertsch2025oolong}.} Oolong-real is a long-context information aggregation benchmark constructed from real conversational transcripts of the \textit{Critical Role} Dungeons and Dragons (DnD) series. The benchmark contains questions over long multi-episode transcripts involving counting, aggregation, indexing, and retrieval tasks across conversational contexts that can span up to millions of tokens. We evaluate on a randomly sampled subset of roughly 500 examples from the original 6,072--example benchmark.

The benchmark contains three answer types: numeric, string, and lists of strings. Scoring follows exact match for string answers and set overlap for list answers. However, the original scoring ${0.75}^{|\hat{y}-y|}$ on numerics leads to undesirable behavior: for example, predicting 1 instead of 10 receives the same score as predicting 991 instead of 1000. It also heavily penalizes small absolute differences at large scales (e.g., 10000 vs. 9985 scores 1\%). To address this, we replace the original numeric scorer with \textit{relaxed accuracy}, where a \emph{numeric} prediction is counted as correct if it lies within 5\% of the gold answer which is standard choice in QA tasks~\citep{DBLP:conf/acl/MasryLTJH22, DBLP:conf/acl/MasryIABKKLRRST25}.

Finally, we use \texttt{gpt-5-nano} to parse model outputs into the expected structured answer format before evaluation.

\paragraph{SynthWorlds (SM) \citep{gu2025synthworlds}.} SynthWorlds-SM evaluates multi-hop reasoning in a synthetic Wikidata environment where entity names are systematically replaced, preventing models from relying on memorized world knowledge. We evaluate on all 1,200 test examples from the synthetic-mapped (SM) split. We report per-token F1 as in the original paper~\citep{gu2025synthworlds}. The benchmark contains multiple reasoning structures (e.g., linear chains, diamonds, and compositional motifs) requiring multi-step aggregation across documents. We provide in-context examples covering all six reasoning paradigms used in the benchmark on all baselines. Finally, the retrieval tool for fetching entity-documents implemented using OpenAI's \texttt{text-embedding-3-small} embeddings with cosine similarity search. Following the original setup~\citep{gu2025synthworlds}, the retriever returns the top-5 documents from the 6k+ document corpus for each query.

\paragraph{PhantomWiki \citep{gong2025phantomwiki}.} PhantomWiki tests multi-hop QA over a synthetic universes of configurable size. Each universe contains entities, relationships, and supporting wiki-style documents, enabling controlled evaluation of multi-hop reasoning and retrieval. We evaluate on a PhantomWiki universe of size 500 generated with seed 1. Following the original work~\citep{gong2025phantomwiki}, we use the same 10 in-context examples provided for the ReAct baseline, adapting them separately for each method to match its expected patterns (e.g., CodeAct writes python \texttt{<code>...</code>} blocks instead of JSON actions). Finally, For we report F1 score computed from the number of exact-match answers shared between the predicted and gold answer sets, as defined in the original paper.

\paragraph{DeepSearchQA \citep{gupta2026deepsearchqa}.} DeepSearchQA evaluates agents on complex, multi-step information-seeking tasks that require discovering, aggregating, and verifying information from the open web. The benchmark emphasizes long-horizon search planning, retrieval strategy, and meta-reasoning about when sufficient information has been collected. We evaluate on the 316-question \textit{single-answer} subset from the full 900-question benchmark. We restrict evaluation to this split because the remaining set-based questions require nuanced partial-credit grading, making reliable automatic evaluation substantially harder. Instead, we use a binary correctness setup (\textit{correct}/\textit{incorrect}), which is significantly more stable for LLM-as-a-judge evaluation. We use \texttt{gpt-5-nano} as the judge model for answer verification. We report accuracy.

\subsection{Computational Resources}
\label{app:compute}

Inference servers were deployed using vLLM v0.20.0 \citep{kwon2023efficient} 
and managed via SLURM on nodes equipped with NVIDIA H200 GPUs. All experiments 
reported in this paper collectively required approximately 350 GPU hours. vLLM server configuration and 
job submission scripts are provided along with our code and configurations at \codelink.

\section{Detailed Results}\label{sec:detailed-results}

In Table~\ref{tab:main-results-error-bars} we showcase the same exact numbers from Table~\ref{tab:main-results} with  standard error of the mean (SEM) bars.

\begin{table}
\centering
\scriptsize
\hspace*{-0.65cm}
\begin{tabular}{l l cccc c}
\cmidrule(lr){2-7}
\multirow{3}{*}{} & \multirow{3}{*}{\diagbox[width=11em]{\textbf{Benchmark}}{\textbf{Method}}}
& \multirow[c]{3}{*}{\raisebox{0.7em}{\textbf{ReAct}}}
& \multirow[c]{3}{*}{\raisebox{0.7em}{\textbf{CodeAct}}}
& \textbf{Deep}
& \multirow[c]{3}{*}{\raisebox{0.7em}{\textbf{RLM}}}
& \multirow[c]{3}{*}{\raisebox{0.7em}{\textbf{\DR{}}}} \\
& & & &\textbf{Research} &  & \\
& 
& \raisebox{0.35em}{\citeyearpar{yao2022react}}
& \raisebox{0.35em}{\citeyearpar{wang2024executable}}
& \raisebox{0.12em}{\citeyearpar{roucher_open_deep_research_2025}}
& \raisebox{0.35em}{\citeyearpar{zhang2025recursive}}
& \raisebox{0.35em}{\textit{(ours)}} \\
\cmidrule(lr){2-7}
\multirow{4}{*}{\rotatebox{90}{\scriptsize\textit{\shortstack{Qwen3 8B \\ Thinking}}}}

& SynthWorlds~\citeyearpar{gu2025synthworlds}
   & 0.176 $\pm$ 0.011 & 0.268 $\pm$ 0.012 & 0.206 $\pm$ 0.011 & 0.058 $\pm$ 0.006 & \textbf{0.305 $\pm$ 0.013} \\
& PhantomWiki \citeyearpar{gong2025phantomwiki}
   & 0.144 $\pm$ 0.014 & 0.153 $\pm$ 0.015 & 0.120 $\pm$ 0.014 & 0.043 $\pm$ 0.008 & \textbf{0.172 $\pm$ 0.016} \\
& DeepSearchQA \citeyearpar{gupta2026deepsearchqa}
   & 0.124 $\pm$ 0.019 & 0.149 $\pm$ 0.020 & 0.142 $\pm$ 0.020 & 0.139 $\pm$ 0.020 & \textbf{0.161 $\pm$ 0.021} \\
& Oolong-\textit{real}~\citeyearpar{bertsch2025oolong}
   & NA$^\dagger$ & 0.045 $\pm$ 0.014 & NA$^\dagger$ & 0.065 $\pm$ 0.015 & \textbf{0.076 $\pm$ 0.011} \\
\cmidrule(lr){2-7}
\multirow{4}{*}{\rotatebox{90}{\scriptsize\textit{\shortstack{Qwen3 32B \\ Thinking}}}}

& SynthWorlds~\citeyearpar{gu2025synthworlds}
   & 0.228 $\pm$ 0.012 & 0.281 $\pm$ 0.013 & 0.275 $\pm$ 0.012 & 0.169 $\pm$ 0.010 & \textbf{0.346 $\pm$ 0.013} \\
& PhantomWiki \citeyearpar{gong2025phantomwiki}
   & 0.167 $\pm$ 0.015 & 0.252 $\pm$ 0.017 & 0.212 $\pm$ 0.017 & 0.160 $\pm$ 0.015 & \textbf{0.369 $\pm$ 0.020} \\
& DeepSearchQA \citeyearpar{gupta2026deepsearchqa}
   & 0.177 $\pm$ 0.022 & 0.213 $\pm$ 0.023 & 0.230 $\pm$ 0.024 & 0.190 $\pm$ 0.022 & \textbf{0.241 $\pm$ 0.024} \\
& Oolong-\textit{real}~\citeyearpar{bertsch2025oolong}
   & NA$^\dagger$ & 0.060 $\pm$ 0.015 & NA$^\dagger$ & 0.077 $\pm$ 0.013 & \textbf{0.132 $\pm$ 0.014} \\
\cmidrule(lr){2-7}
\multirow{4}{*}{\rotatebox{90}{\scriptsize\textit{\shortstack{Llama-3.3 \\ 70B Instruct}}}}

& SynthWorlds~\citeyearpar{gu2025synthworlds}
   & 0.300 $\pm$ 0.013 & \textbf{0.480 $\pm$ 0.014} & 0.308 $\pm$ 0.013 & 0.401 $\pm$ 0.014 & 0.359 $\pm$ 0.014 \\
& PhantomWiki \citeyearpar{gong2025phantomwiki}
   & 0.244 $\pm$ 0.017 & 0.381 $\pm$ 0.020 & 0.160 $\pm$ 0.015 & 0.270 $\pm$ 0.018 & \textbf{0.512 $\pm$ 0.021} \\
& DeepSearchQA~\citeyearpar{gupta2026deepsearchqa}
   & 0.155 $\pm$ 0.020 & 0.184 $\pm$ 0.022 & 0.127 $\pm$ 0.019 & 0.175 $\pm$ 0.021 & \textbf{0.187 $\pm$ 0.022} \\
& Oolong-\textit{real}~\citeyearpar{bertsch2025oolong}
   & NA$^\dagger$ & 0.079 $\pm$ 0.012 & NA$^\dagger$ & 0.064 $\pm$ 0.012 & \textbf{0.151 $\pm$ 0.015} \\
\cmidrule(lr){2-7} \\
\end{tabular}
\vspace{1mm}
\caption{Scores (0--1, $\uparrow$ higher is better) across four reasoning benchmarks. NA$^\dagger$ indicates context size limitations. }
% \mt{remove surrounding lines}
\label{tab:main-results-error-bars}
\end{table}

\section{Failure analysis}\label{apn:failure_mode}

We analyze failure modes on the 2,166 tasks where \DR{}  answers correctly and at least one baseline does not, using Qwen3-32B traces across all four baselines and benchmarks. \DR{} succeeds on these tasks largely because it decomposes each problem into recursive $\IDeep$ calls, so that each sub-call independently searches and retrieves the relevant information rather than abandoning the task when a single query fails. Additionally, by routing all quantitative and retrieval operations through $\Ecal$, each LLM call produces a small formal model rather than a complete answer, reducing the surface area for hallucinations.

For each failing baseline trace we construct a normalized view that preserves the agent's reasoning while compressing tool outputs and repeated content. Following the LLM-as-judge paradigm~\citep{zheng2023judging}, an LLM judge (Gemini-3-Flash \footnote{\href{https://deepmind.google/models/gemini/flash/}{https://deepmind.google/models/gemini/flash/}}) receives this trace along with the task question, gold answer, and the agent's final answer, and produces a structured output of a narrative summary, an open-coded primary and optional secondary failure mode, and a verbatim evidence quote from the trace. 

To derive reportable statistics from these open-coded phrases, we first cluster all outputs using HDBSCAN over BGE-large \citep{chen2024m3} embeddings to discover natural groupings, following the embedding clusters and naming pipeline of~\citet{tamkin2024clio}. This process results in a fixed multi-label taxonomy of seven categories: (1)premature termination, (2) hallucination/fabrication, (3) incomplete search traversal, (4) unreliable delegation/aggregation, (5) reasoning/logic error, (6) tool misuse/scaffold error, and (7) other. A second LLM pass (Gemini-3-Flash) classifies each example into this taxonomy, assigning all applicable labels. A single example can have multiple labels. For instance, an agent that hallucinates an answer and then terminates receives both premature termination and hallucination.

To validate the automated classifications, three PhD  annotators independently reviewed a stratified sample of 50 judge outputs covering all baselines and taxonomy categories. Each annotator saw the judge's summary, evidence quote, and open-coded failure modes alongside the classifier's labels, and marked each label as correct or incorrect with the option to add missed labels. Mean agreement with LLM-judge classifications was 0.92 amongst the 3 annotators, with pairwise Inter-Annotator Agreement (bias adjusted kappa \citep{byrt1993bias}) of 0.82, overall indicating strong agreement with the automated pipeline.

\section{Qualitative Trace Analysis}\label{apn:qualitative_examples}

We present three side-by-side trace comparisons from the \texttt{dr\_only} cell (\DR{} correct, baseline incorrect, Qwen3-32B).  These examples were selected to illustrate the most prevalent failure modes identified by our taxonomy, namely, premature termination, and hallucination/fabrication, which together account for the majority of baseline failures.

% \dl{TODO make the example task and gold answer their own colorbox}
% \dl{TODO Try to make claude use the macros i used to make these colored and more readable, if it gets complicated, dont sweat it.}
% \dl{Say something about percentage of failure modes that where in these clusters if it makes sense, otherwise stress that this are indicative examples that you selected to show the behaviors}
% Example 1: CodeAct vs DR on SynthWorlds ────────────────────────────────
% \kg{correct colors}
\noindent\textbf{Example 1: CodeAct hallucinates on (SynthWorlds, \texttt{0068-qa-sm})}.

\begin{tcolorbox}[taskbox]
What is the opposite of the script in which ``Flames of Destiny'' is written?\\
\textbf{Gold answer:} Valtaroscript
\end{tcolorbox}

\vspace{4pt}
\noindent\textbf{CodeAct} (score: 0, answer: ``Alphabetic Script''):

\begin{tcolorbox}[colback=black!3, colframe=black!20, fontupper=\ttfamily\scriptsize, boxrule=0.4pt, left=2pt, right=2pt, top=2pt, bottom=2pt, breakable]
Thought: I need to determine the script format of ``Flames of Destiny''
to find its opposite.\\
\textless code\textgreater\\
docs = search(query="Flames of Destiny script")\\
print(docs)\\
\textless/code\textgreater\\[4pt]
Observation:\\
{[}'Flames of Destiny is a video game released in 2057. ... The game
includes a wide set of languages and scripts for its in-game text ...
Among the scripts used are Valtarian Glyphic Script (a writing system
described as a script), ...'{]}\\[4pt]
Thought: The script used in ``Flames of Destiny'' is the Valtarian
Glyphic Script. The opposite of a glyphic (logographic) script is
typically an alphabetic script. Since the game supports multiple
languages like Kynthor and Velthar (alphabetic), the opposite is an
alphabetic script.\\
\textless code\textgreater\\
final\_answer("Alphabetic Script")\\
\textless/code\textgreater
\end{tcolorbox}

\noindent CodeAct identifies ``Valtarian Glyphic Script'' correctly but never issues a second search. Instead, it hallucinates a real-world typological contrast (``glyphic $\to$ alphabetic'') from parametric knowledge.

\vspace{6pt}
\noindent\textbf{\DR} (score: 1, answer: ``Valtaroscript''):

\begin{repl}
docs = search("Flames of Destiny")
script = llm(f"What script is 'Flames of Destiny' written in? "
             f"Return just the script name.\n\n"
             f"{''.join(docs)}")
print(script)
\end{repl}
\begin{observation}
Valtarian Glyphic Script
\end{observation}
\begin{repl}
docs = search("Valtarian Glyphic Script")
opposite_script = llm(f"What is the opposite of the Valtarian Glyphic "
                       f"Script? Return just the script name.\n\n"
                       f"{''.join(docs)}")
print(opposite_script)
\end{repl}
\begin{observation}
Valtaroscript
\end{observation}
\begin{repl}
FinalAnswer("Valtaroscript")
\end{repl}

\noindent Each hop is a separate search $+$ extraction call (\textbf{P2}). The second sub-LLM receives only the documents about \emph{Valtarian Glyphic Script} and can read the in-universe ``opposite'' relation directly---no parametric guessing required.

%% ─── Example 2: ReAct vs DR on PhantomWiki ──────────────────────────────────
\noindent\textbf{Example 2: ReAct terminates prematurely on implicit relation (PhantomWiki-500, \texttt{2f4ee116}).}
\\
\begin{tcolorbox}[taskbox]
Who is the friend of the daughter-in-law of the person whose date of birth is 0984-05-03?\\
\textbf{Gold answer:} Bobbie Luu
\end{tcolorbox}

\vspace{4pt}
\noindent\textbf{ReAct} (score: 0, answer: empty):

\begin{tcolorbox}[colback=black!3, colframe=black!20, fontupper=\ttfamily\scriptsize, boxrule=0.4pt, left=2pt, right=2pt, top=2pt, bottom=2pt, breakable]
Tool call: search(attribute="0984-05-03")\\
Observation: (1) Earle Coe\\[4pt]
Tool call: retrieve\_article(entity="Earle Coe")\\
Observation:\\
\# Earle Coe\\
\#\# Family\\
The mother of Earle Coe is Alycia Coe.\\
The father of Earle Coe is Christoper Coe.\\
The son of Earle Coe is Reggie Coe.\\
The wife of Earle Coe is Christina Coe.\\
\#\# Attributes\\
The date of birth of Earle Coe is 0984-05-03.\\
The occupation of Earle Coe is petroleum engineer. ...\\[4pt]
Action: \{"name": "final\_answer", "arguments": \{"answer": ""\}\}
\end{tcolorbox}

\noindent ReAct finds Earle Coe and sees no explicit ``daughter-in-law'' field. It immediately gives up---never looking up the son's wife.

\vspace{6pt}
\noindent\textbf{\DR} (score: 1, answer: ``Bobbie Luu''):

\begin{repl}
search_result = search("0984-05-03")
print(f"Search result for DOB: {search_result}")
\end{repl}
\begin{observation}
Search result for DOB: (1) Earle Coe
\end{observation}
\begin{repl}
article = retrieve_article("Earle Coe")
print(f"Earle Coe's article: {article}")
\end{repl}
\begin{observation}
... The son of Earle Coe is Reggie Coe. The wife of Earle Coe is Christina Coe. ...
\end{observation}

\begin{repl}
# No ``daughter-in-law'' field; check if son has a wife
reggie_article = retrieve_article("Reggie Coe")
print(f"Reggie Coe's article: {reggie_article}")
\end{repl}
\begin{observation}
... The wife of Reggie Coe is Lissa Coe. ...
\end{observation}
\begin{repl}
lissa_article = retrieve_article("Lissa Coe")
print(f"Lissa Coe's article: {lissa_article}")
\end{repl}
\begin{observation}
... The friend of Lissa Coe is Bobbie Luu. ...
\end{observation}
\begin{repl}
FinalAnswer(["Bobbie Luu"])
\end{repl}

\noindent \DR{} reasons incrementally (\textbf{P3}): seeing no daughter-in-law field, it infers the relation must be resolved via son $\to$ wife and issues the next retrieval. Each hop is a separate \texttt{<repl>} turn (\textbf{P2}), making continuation a consequence of the code's control flow rather than an optional associative decision.

%% ─── Example 3: RLM vs DR on Oolong-real ────────────────────────────────────
\noindent\textbf{Example 3: RLM uses brittle heuristics for counting (Oolong-real, \texttt{18ac05eb})}

\begin{tcolorbox}[taskbox]
Across all episodes, what percentage of rolls were of value 4? Round to the nearest integer.\\
\textbf{Gold answer:} 2
\end{tcolorbox}

\vspace{4pt}
\noindent\textbf{RLM} (score: 0, answer: 8):

\begin{tcolorbox}[colback=black!3, colframe=black!20, fontupper=\ttfamily\scriptsize, boxrule=0.4pt, left=2pt, right=2pt, top=2pt, bottom=2pt, breakable]
```repl\\
import csv\\
from io import StringIO\\
csv\_data = StringIO(context)\\
csv\_reader = csv.DictReader(csv\_data)\\
count\_4 = 0; total\_rolls = 0\\
for row in csv\_reader:\\
\hspace*{1em}try:\\
\hspace*{2em}roll\_value = int(row.get('roll\_value', 0))\\
\hspace*{2em}total\_rolls += 1\\
\hspace*{2em}if roll\_value == 4: count\_4 += 1\\
\hspace*{1em}except ValueError: continue\\
```\\[4pt]
\# CSV parsing fails on unstructured dialogue; try regex\\
```repl\\
import re\\
roll\_pattern = re.compile(r'\textbackslash b4\textbackslash b')\\
count\_4 = len(roll\_pattern.findall(context))\\
total\_pattern = re.compile(r'\textbackslash b\textbackslash d\textbackslash b')\\
total\_rolls = len(total\_pattern.findall(context))\\
percentage = (count\_4 / total\_rolls * 100)\\
print(f"Regex-based estimate: \{round(percentage)\}\%")\\
```\\[4pt]
... [4 more iterations with different regex heuristics] ...\\[4pt]
```repl\\
FINAL\_VAR(final\_percentage)  \# outputs 8\\
```
\end{tcolorbox}

\noindent RLM receives the entire multi-episode transcript in a single \texttt{context} variable. It tries heuristics that are not reliably able to identify dice rolls in unstructured D\&D dialogue. The final answer (8) is a spurious approximation from counting standalone digit ``4'' occurrences across all text.
\vspace{6pt}
% \dl{Across all of our traces, change plan\_exec and stuff like that to DeepReasoner. You can keep the add\_task and run\_all behavior}

\noindent\textbf{\DR} (score: 1, answer: 2):

\begin{repl}
episodes = DoLoReS(
    """Extract all episodes from the DnD game text in `document`.
    Each episode is delimited by '[START OF EPISODE]' and
    '[END OF EPISODE]'. Return a list of strings.""",
    document=document, namespace="splitting episodes")
print(f"Found {len(episodes)} episodes.")
\end{repl}
\begin{observation}
Found 8 episodes.
\end{observation}
\begin{repl}
for ep in episodes:
    DoLoReS.add_task(
        """In the given DnD episode text, count the total number
        of dice rolls and how many resulted in a value of 4.
        Return a tuple (total_rolls, fours_count) as integers.""",
        episode=Var(ep, "DnD episode text"),
        namespace="episode qa")
results = DoLoReS.run_all()

total_rolls = sum(r[0] for r in results)
total_fours = sum(r[1] for r in results)
percentage = round((total_fours / total_rolls) * 100)
print(f"Total rolls: {total_rolls}, Total fours: {total_fours}, "
      f"Percentage: {percentage}%")
\end{repl}
\begin{observation}
Total rolls: 1188, Total fours: 26, Percentage: 2%
\end{observation}
\begin{repl}
FinalAnswer("2")
\end{repl}

\noindent \DR{} splits the document into episodes via delimiter matching (\textbf{P1}), dispatches 8 parallel sub-agents that each count within a single bounded episode (\textbf{P2}), and aggregates with Python arithmetic (\textbf{P1}). Each sub-LLM handles a tractable chunk rather than the full 100k+ token transcript.

\section{Ablation: In-context decomposition examples}
\label{apn:ablation-examples}

Table~\ref{tab:ablation-examples} reports the full results of the in-context decomposition ablation discussed in main text. We compare the full \DR{} agent against two in context variants on Qwen3-32B: \textit{no-examples}, which removes all decomposition examples from the system prompt, and \textit{with-principles}, which replaces the examples with natural-language descriptions of the three design principles from \S\ref{sec:design-principles}. Across all four benchmarks, removing the in-context examples leads to substantial drops in both F1 and exact match, and replacing them with natural-language principles is uniformly worse in F1 than removing them altogether.

\begin{table}[h]
\centering
\small
\begin{tabular}{llcc}
\toprule
Benchmark & Variant & F1 & EM \\
\midrule
\multirow{3}{*}{SynthWorlds~\citeyearpar{gu2025synthworlds}}
 & \textit{no-examples}     & 0.041 & 0.006 \\
 & \textit{with-principles} & 0.036 & 0.007 \\
 & \DR{}                    & \textbf{0.345} & \textbf{0.325} \\
\midrule
\multirow{3}{*}{PhantomWiki~\citeyearpar{gong2025phantomwiki}}
 & \textit{no-examples}     & 0.113 & 0.076 \\
 & \textit{with-principles} & 0.080 & 0.052 \\
 & \DR{}                    & \textbf{0.373} & \textbf{0.279} \\
\midrule
\multirow{3}{*}{DeepSearchQA~\citeyearpar{gupta2026deepsearchqa}}
 & \textit{no-examples}     & 0.139 & 0.139 \\
 & \textit{with-principles} & 0.117 & 0.117 \\
 & \DR{}                    & \textbf{0.241} & \textbf{0.241} \\
\midrule
\multirow{3}{*}{Oolong-\textit{real}~\citeyearpar{bertsch2025oolong}}
 & \textit{no-examples}     & 0.036 & 0.033 \\
 & \textit{with-principles} & 0.033 & 0.031 \\
 & \DR{}                    & \textbf{0.137} & \textbf{0.107} \\
\bottomrule
\end{tabular}
\vspace{3mm}
\caption{Ablation of \DR{}'s in-context decomposition examples on Qwen3-32B. \textit{no-examples} removes the atomic decomposition examples from the system prompt; \textit{with-principles} replaces them with natural-language statements of the design principles from \S\ref{sec:design-principles}. Both reduced variants degrade sharply against the full \DR{} agent on every benchmark, and \textit{with-principles} is consistently worse than \textit{no-examples} in F1, suggesting that current LLMs operationalize structured meta-reasoning by analogy to concrete examples rather than by following stated rules.}
\label{tab:ablation-examples}
\end{table}

\section{Token-Matched Baselines}
\label{apn:token-matched}

\DR{} spawns many reasoning threads and therefore consumes more tokens per task than the baselines. To test whether the baselines benefit from a comparable budget, we evaluate a token-matched ensemble of the strongest baseline on each benchmark using Qwen3-32B. we compute the average number of tokens per task used by \DR{} and by the strongest baseline, and set $K$ to their ratio, rounded up, for each benchmark. We then run the baseline $K$ times on every task and select the final answer by majority vote over the $K$ runs, using an LLM judge to group equivalent answers before voting. Scores below use the metrics described in \S\ref{apn:exp_details}.

\begin{table}[h]
\centering
\small
\begin{tabular}{llcccc}
\toprule
Benchmark & Baseline & $K$ & Single run & Majority@$K$ & \DR{} \\
\midrule
SynthWorlds~\citeyearpar{gu2025synthworlds}      & CodeAct       & 3  & 0.313 & 0.337 & \textbf{0.346} \\
PhantomWiki~\citeyearpar{gong2025phantomwiki}    & CodeAct       & 10 & 0.266 & 0.247 & \textbf{0.369} \\
DeepSearchQA~\citeyearpar{gupta2026deepsearchqa} & Deep Research & 3  & 0.155 & 0.159 & \textbf{0.241} \\
Oolong-\textit{real}~\citeyearpar{bertsch2025oolong} & RLM       & 4  & 0.072 & 0.072 & \textbf{0.132} \\
\bottomrule
\end{tabular}
\vspace{3mm}
\caption{Token-matched evaluation on Qwen3-32B. $K$ is the ratio of \DR{}'s average token usage to that of the strongest baseline, rounded up. Majority@$K$ reports majority voting over $K$ independent baseline runs per task. \DR{} outperforms the token-matched baseline on every benchmark.}
\label{tab:token-matched}
\end{table}

Table~\ref{tab:token-matched} shows that majority voting changes baseline scores by between $-0.019$ and $+0.024$, and \DR{} outperforms the token-matched baseline on all four benchmarks. This is consistent with the failure analysis in \S\ref{apn:failure_mode}: when individual trajectories terminate prematurely or hallucinate, independent runs tend to fail in similar ways, and voting over them rarely recovers the correct answer.

Single-run scores differ from Table~\ref{tab:main-results} because the baselines were re-run for this experiment. Deep Research on DeepSearchQA was unstable across runs (0.155 here vs.\ 0.230 in Table~\ref{tab:main-results}). For unstable baselines, to be as strict as possible, we report the highest-scoring run. We did not observe comparable instability for \DR{}, which also outperforms the Table~\ref{tab:main-results} scores of all baselines.

\section{Recursion Depth of \DR{}}\label{app:recursion-depth}

\begin{table}[t]
\centering
\begin{tabular}{lrrrr}
\toprule
Model & Runs & Mean & \#$>$10 & $\leq$10 (\%) \\
\midrule
Llama-3.3-70B-Instruct & 4{,}716  & 2.88 & 4  & 99.92 \\
Qwen3-32B              & 4{,}716  & 2.95 & 0  & 100.00 \\
Qwen3-8B               & 4{,}716  & 2.82 & 8  & 99.83 \\
\midrule
All                    & 14{,}148 & 2.88 & 12 & 99.92 \\
\bottomrule
\end{tabular}
\vspace{3mm}
\caption{Maximum recursion depth of \textsc{Dolores} across all benchmarks.
Recursion is consistently shallow; fewer than $0.1\%$ of runs exceed depth 10.}
\label{tab:recursion-depth}
\end{table}

\paragraph{Recursion Depth is Well-Behaved.} Throughout our experiments we closely monitored the recursion depth, since unbounded recursion could in principle lead to non-termination. As shown in Table~\ref{tab:recursion-depth}, across all 14,148 runs (all models and benchmarks), the recursion depth remains shallow: the mean maximum recursion depth is 2.9, and more than 99.9\% of runs terminate at a maximum depth of 10 or below. The behaviour is stable across models---Qwen3-32B never exceeds depth 7---and only 12 of 14,148 runs (0.08\%) exceed depth 10, all of which terminated.

\paragraph{Deployment Considerations} In order to deploy \DR{} in production, it is advisable to enforce an explicit maximum recursion depth (e.g., 10). This bounds the worst case while leaving more than 99.9\% of runs unaffected, as shown in our experiments Table~\ref{tab:recursion-depth}.

\clearpage

\ifDRchecklist
\section*{NeurIPS Paper Checklist}

\begin{enumerate}

\item {\bf Claims}
    \item[] Question: Do the main claims made in the abstract and introduction accurately reflect the paper's contributions and scope?
    \item[] Answer: \answerYes{} % Replace by \answerYes{}, \answerNo{}, or \answerNA{}.
    \item[] Justification: The abstract and introduction state the main contribution, scope, and empirical claims.

\item {\bf Limitations}
    \item[] Question: Does the paper discuss the limitations of the work performed by the authors?
    \item[] Answer: \answerYes{} % Replace by \answerYes{}, \answerNo{}, or \answerNA{}.
    \item[] Justification: The paper discusses its limitations in \S\ref{sec:limitations}.

\item {\bf Theory assumptions and proofs}
    \item[] Question: For each theoretical result, does the paper provide the full set of assumptions and a complete (and correct) proof?
    \item[] Answer: \answerNA{}% Replace by \answerYes{}, \answerNo{}, or \answerNA{}.
    \item[] Justification: The paper does not contain theoretical results requiring formal proofs.

    \item {\bf Experimental result reproducibility}
    \item[] Question: Does the paper fully disclose all the information needed to reproduce the main experimental results of the paper to the extent that it affects the main claims and/or conclusions of the paper (regardless of whether the code and data are provided or not)?
    \item[] Answer: \answerYes{} % Replace by \answerYes{}, \answerNo{}, or \answerNA{}.
    \item[] Justification: The information required to reproduce our work is provided in \S\ref{sec:deep_reasoning}, \S\ref{sec:experiments} and Appendix~\ref{apn:exp_details}.

\item {\bf Open access to data and code}
    \item[] Question: Does the paper provide open access to the data and code, with sufficient instructions to faithfully reproduce the main experimental results, as described in supplemental material?
    \item[] Answer: \answerYes{} % Replace by \answerYes{}, \answerNo{}, or \answerNA{}.
    \item[] Justification: We provide open access to data and code at \codelink.

\item {\bf Experimental setting/details}
    \item[] Question: Does the paper specify all the training and test details (e.g., data splits, hyperparameters, how they were chosen, type of optimizer) necessary to understand the results?
    \item[] Answer: \answerYes{} % Replace by \answerYes{}, \answerNo{}, or \answerNA{}.
    \item[] Justification: The information required to understand the results of our work is provided in \S\ref{sec:deep_reasoning}, \S\ref{sec:experiments} and Appendix~\ref{apn:exp_details}.
    

\item {\bf Experiment statistical significance}
    \item[] Question: Does the paper report error bars suitably and correctly defined or other appropriate information about the statistical significance of the experiments?
    \item[] Answer: \answerYes{} % Replace by \answerYes{}, \answerNo{}, or \answerNA{}.
    \item[] Justification: We report the standard error of the mean on all experiments in Appendix~\ref{sec:detailed-results}.

\item {\bf Experiments compute resources}
    \item[] Question: For each experiment, does the paper provide sufficient information on the computer resources (type of compute workers, memory, time of execution) needed to reproduce the experiments?
    \item[] Answer: \answerYes{} % Replace by \answerYes{}, \answerNo{}, or \answerNA{}.
    \item[] Justification:  We provide information on compute resources needed to reproduce experiments in Appendix~\ref{app:compute}.
    
\item {\bf Code of ethics}
    \item[] Question: Does the research conducted in the paper conform, in every respect, with the NeurIPS Code of Ethics \url{https://neurips.cc/public/EthicsGuidelines}?
    \item[] Answer: \answerYes{} % Replace by \answerYes{}, \answerNo{}, or \answerNA{}.
    \item[] Justification: We read and conform to the NeurIPS Code of Ethics.

\item {\bf Broader impacts}
    \item[] Question: Does the paper discuss both potential positive societal impacts and negative societal impacts of the work performed?
    \item[] Answer: \answerYes{} % Replace by \answerYes{}, \answerNo{}, or \answerNA{}.
    \item[] Justification:  We discuss the limitations of the work in \S\ref{sec:limitations} and its potential positive and negative societal impacts in \S\ref{sec:impact}.
    
\item {\bf Safeguards}
    \item[] Question: Does the paper describe safeguards that have been put in place for responsible release of data or models that have a high risk for misuse (e.g., pre-trained language models, image generators, or scraped datasets)?
    \item[] Answer: \answerNA{}{} % Replace by \answerYes{}, \answerNo{}, or \answerNA{}.
    \item[] Justification: The paper poses no such risks.

\item {\bf Licenses for existing assets}
    \item[] Question: Are the creators or original owners of assets (e.g., code, data, models), used in the paper, properly credited and are the license and terms of use explicitly mentioned and properly respected?
    \item[] Answer: \answerYes{}{} % Replace by \answerYes{}, \answerNo{}, or \answerNA{}.
    \item[] Justification: We cite the related assets and follow their license and terms.

\item {\bf New assets}
    \item[] Question: Are new assets introduced in the paper well documented and is the documentation provided alongside the assets?
    \item[] Answer: \answerYes{} % Replace by \answerYes{}, \answerNo{}, or \answerNA{}.
    \item[] Justification: We include our code at \codelink.

\item {\bf Crowdsourcing and research with human subjects}
    \item[] Question: For crowdsourcing experiments and research with human subjects, does the paper include the full text of instructions given to participants and screenshots, if applicable, as well as details about compensation (if any)? 
    \item[] Answer: \answerNA{} % Replace by \answerYes{}, \answerNo{}, or \answerNA{}.
    \item[] Justification: Only members of the research team were involved in human validation.

\item {\bf Institutional review board (IRB) approvals or equivalent for research with human subjects}
    \item[] Question: Does the paper describe potential risks incurred by study participants, whether such risks were disclosed to the subjects, and whether Institutional Review Board (IRB) approvals (or an equivalent approval/review based on the requirements of your country or institution) were obtained?
    \item[] Answer: \answerNA{} % Replace by \answerYes{}, \answerNo{}, or \answerNA{}.
    \item[] Justification:  Only members of the research team were involved in human validation.

\item {\bf Declaration of LLM usage}
    \item[] Question: Does the paper describe the usage of LLMs if it is an important, original, or non-standard component of the core methods in this research? Note that if the LLM is used only for writing, editing, or formatting purposes and does \emph{not} impact the core methodology, scientific rigor, or originality of the research, declaration is not required.
    %this research? 
    \item[] Answer: \answerYes{} % Replace by \answerYes{}, \answerNo{}, or \answerNA{}.
    \item[] Justification: LLMs are the core object of study: \DR{} is an agentic scaffold built from LLM calls, described in \S\ref{sec:deep_reasoning} and Appendix~\ref{apn:prompts}. We additionally use LLM judges as measurement instruments, for answer verification (Appendix~\ref{apn:exp_details}) and for the failure-mode analysis (Appendix~\ref{apn:failure_mode}), the latter validated against human annotation as reported there.

\end{enumerate}
\fi % \ifDRchecklist

\end{document}